\documentclass[journal]{IEEEtran}

\usepackage{graphicx}
\usepackage{bm}
\usepackage{subfigure}
\usepackage{algorithm}
\usepackage{algpseudocode}
\usepackage{amsmath}
\usepackage{multirow}
\usepackage{multicol}
\usepackage{cite}
\usepackage{threeparttable}
\usepackage{color}
\usepackage{doi}
\usepackage{enumerate}
\usepackage{caption}
\usepackage{rotating}
\usepackage{multirow}
\usepackage{threeparttable}
\usepackage{booktabs}
\usepackage{makecell}
\usepackage{setspace}
\usepackage[square, comma, sort&compress, numbers]{natbib}

\begin{document}
\title{Multi-factorial Optimization for Large-scale Virtual Machine Placement in Cloud Computing}
\author{Zhengping Liang, Jian Zhang, Liang Feng, Zexuan Zhu\IEEEmembership{}

\thanks{Manuscript received XX. XX. XXX; revised XX. XX. XXX; accepted XX. XX. XXX. This work was supported in part by the National Natural Science Foundation of China (Grant no.61471246 and 61672358).\par
Z. Liang, J. Zhang and Z. Zhu are with the College of Computer Science and Software Engineering, Shenzhen University, Shenzhen 518060, China (e-mail: liangzp@szu.edu.cn; roykin1002@163.com; zhuzx@szu.edu.cn).

L. Feng is with the College of Computer Science, Chongqing University, Chongqing, China(e-mail: liangf@cqu.edu.cn).
}}

\maketitle
		
\begin{abstract}
The placement scheme of virtual machines (VMs) to physical servers (PSs) is crucial to lowering operational cost for cloud providers. Evolutionary algorithms (EAs) have been performed promising-solving on virtual machine placement (VMP) problems in the past. However, as growing demand for cloud services, the existing EAs fail to implement in large-scale virtual machine placement (LVMP) problem due to the high time complexity and poor scalability. Recently, the multi-factorial optimization (MFO) technology has surfaced as a new search paradigm in evolutionary computing. It offers the ability to evolve multiple optimization tasks simultaneously during the evolutionary process. This paper aims to apply the MFO technology to the LVMP problem in heterogeneous environment. Firstly, we formulate a deployment cost based VMP problem in the form of the MFO problem. Then, a multi-factorial evolutionary algorithm (MFEA) embedded with greedy-based allocation operator is developed to address the established MFO problem. After that, a re-migration and merge operator is designed to offer the integrated solution of the LVMP problem from the solutions of MFO problem. To assess the effectiveness of our proposed method, the simulation experiments are carried on large-scale and extra large-scale VMs test data sets. The results show that compared with various heuristic methods, our method could shorten optimization time significantly and offer a competitive placement solution for the LVMP problem in heterogeneous environment.

\end{abstract}
		
\IEEEpeerreviewmaketitle
		
\section{Introduction}
		
\IEEEPARstart {A}{s} a new resource provisioning and computing model, cloud computing makes it easier and more convenient for users to obtain configurable resources demand in an online manner \cite{1}\cite{27}. It not only lowers the online deployment threshold and management complexity but also brings the benefits of reliability and low-risk \cite{2}\cite{28}. With the gradual maturity of cloud computing technology, the demand for cloud service by the small and large scale industries are continuously rising, which lead to the expansion of the data center together \cite{3}. The choice of virtual machine placement (VMP) strategy is vital to lower operational cost for cloud service providers.

Virtual machines (VMs) are created according to the hardware computing resources demand of tenants, i.e., CPU, RAM and disk. They independently run an operating system and some applications \cite{4}. The virtualization technology lies the key to the success of cloud computing, which allows multiple VMs to run on the same physical server (PS) simultaneously without mutual interruption on each other \cite{5}. The VMP problems can be given as to seek an optimal solution for placing VMs onto PSs \cite{6}\cite{7}. Since the specified VMP problems can refer to NP-Hard problems, it poses challenges for the researchers in finding an optimal solution \cite{8}\cite{9}.

Up till now, a series of VMP strategies, including heuristic and mathematical methods, have been investigated from different perspectives. According to the difference of target optimization objective, existing researches can be loosely classified as follows: i) target for energy efficient optimization, e.g., an energy-efficient adaptive resource scheduler for networked fog centers (NetFCs) \cite{14}, a minimum energy VM scheduling algorithm (MinES) \cite{15}, a holistic virtual machine scheduling algorithm (GRANITE) \cite{32}, an energy-efficient knee point-driven evolutionary algorithm (EEKnEA) \cite{59}; ii) target for network traffic optimization, e.g., a reliable placement based on the multi-optimization with traffic-aware algorithm (RPMOTA) \cite{16}, a traffic-aware VM optimization (TAVO) scheme on nonuniform memory access (NUMA) systems \cite{17}, a multi-objective ACS algorithm (ACS-BVMP) \cite{18}; iii) target for resource allocation optimization, e.g., a correlation-aware VMP scheme \cite{20}, a layered progressive resource allocation algorithm \cite{29}, an energy-aware resource allocation method (EnReal) \cite{22}, a two-tiered on-demand resource allocation mechanism \cite{26}. Although a serious of methods have been successfully developed to address the VMP problems in cloud computing, most of them are carried out on small-scale VMs simulation experiments.

Evolutionary algorithms (EAs) are classified as one of the population-based heuristic methods that simulating the evolutionary process of the population with optimization characteristics \cite{23}. Owing to the effective global population-based search, EAs have been shown their problem-solving in VMP problems \cite{63}\cite{33}\cite{31}. But the limitation of them is that they often require a large number of fitness evaluations and high time computation to obtain promising solution \cite{35}\cite{36}. As the number of VMs increases in the data center, the VMP problem becomes more complicated and computational expensive. Conventional EAs, which are typically designed to solve a single optimization task, have few practical values on solving the large-scale virtual machine placement (LVMP) problem in the actual cloud environment.

Noted that many real-world optimization tasks tend to be potentially correlated \cite{38}\cite{39}, and some useful knowledge learning from one task could be applied to learn another related task efficiently \cite{40}\cite{41}. Motivated by multi-task learning \cite{43}\cite{42}, the multi-factorial optimization (MFO) technology was introduced to evolve multiple tasks simultaneously by sharing good traits among them in the field of evolutionary computing \cite{45}\cite{58}. Principally, MFO offers the ability to fully exploit the implicit parallelism of population-based search during the evolutionary process \cite{46}\cite{47}.

\begin{figure*}
\captionsetup{font={footnotesize}}
\centering
\includegraphics[scale=0.5]{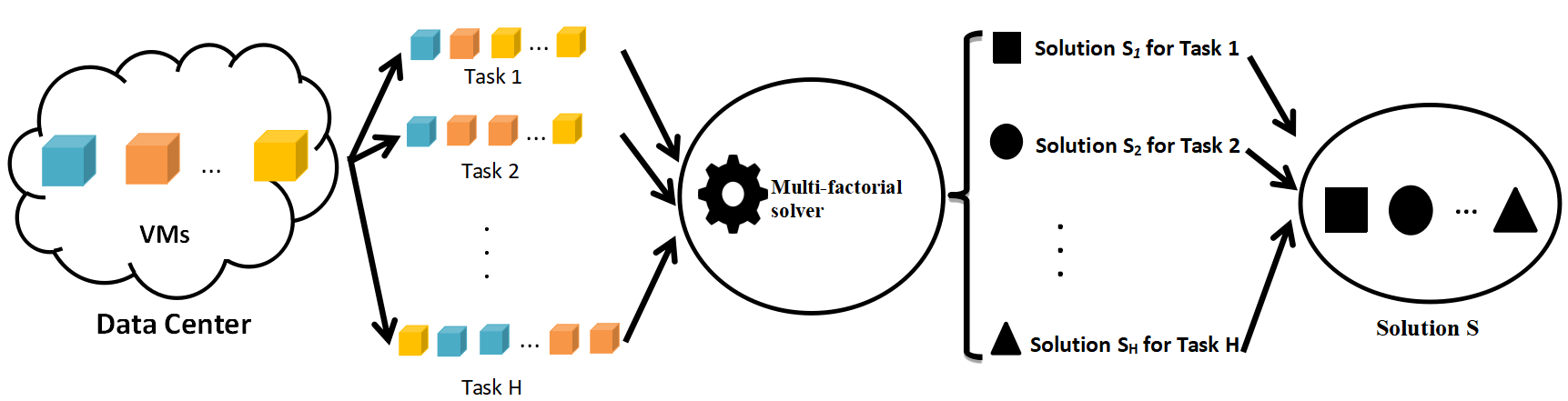}
\captionsetup{justification=centering}
\caption{The core idea of solving the LVMP problem used MFO technology}
\label{f2}
\end{figure*}

This paper aims to develop a new EA-based MFO method to complete the placement of VMs onto PSs from the optimization target of resource allocation. The core idea of this paper is depicted in Fig. \ref{f2} and described as follow. When optimizing the LVMP problem that considered the population-based methods, the population have to contain the information of all VMs in the data center. Nevertheless, it is worth to note that the VMs do not interfere with each other in the data center. Based on this observation, a crucial inspiration is that a single representation of individual in population is actually not desired to preserve the information of all VMs in the data centers. Further, it means that the LVMP problem could be potentially broken down into multiple small-scale virtual machine placement (SVMP) problems, which could be seen as the form of MFO problem. Therefore, the MFO technology is naturally employed to solve multiple SVMP problems within single population simultaneously. As shown in Fig. \ref{f2}, after the iteration process, the solution of the LVMP problem can be gained from the obtained solutions of the established MFO problem. Based on the description above, the main contributions of this paper are summarized as follows:

i). The deployment cost based VMP problem is formulated as the form of the MFO problem in heterogeneous environments of the large-scale data center.

ii). A multi-factorial evolutionary algorithms (MFEA) coupled with greedy-based allocation operator is proposed to solve the established MFO problem.

iii). A re-migration and merge operator is designed to provide the integrated solution of LVMP problem from the solutions of MFO problem.

The rest of this paper is organized as follows. The related work of EA-based VMP methods and the foundation knowledge of MFEA are reviewed in Section II. Section III describes the formulation of VMP problem and the overall framework of the proposed method is provided in Section IV. Section V discuss the simulation experiments and analysis of our proposed method. Finally, Section VI deals with conclusion and future work.

\section{BACKGROUND}
\subsection{EA-based VMP Methods}

Different from the mathematical optimization methods, EA-based methods find the optimal solution through population iteration search without mathematical analysis of the target objective function. In recent years, a series of EA-based methods, such as particle swarm optimization (PSO), genetic algorithm (GA), ant colony system (ACS) and so on, have been developed to address various VMP problems in cloud computing, which are reviewed as follows:

i) \emph{PSO-based VMP methods}. A.-P. Xiong \textit{et al.} \cite{63} designed energy efficient VMs allocation problem using PSO technology. In their study, the fitness function of PSO is defined as the total Euclidean distance to determine the optimal point between resource utilization and energy consumption. But its limitation of this work is that authors considered single type of VMs. P. Aruna \textit{et al.} \cite{65} explored a PSO algorithm for the VM provisioning to make the cloud data centers as power efficient, which involves the discussion of the power model for the servers and the implement of the proposed power aware PSO algorithm for the VM provisioning. Recently, another modified binary PSO is proposed by A. Tripathi \textit{et al.} \cite{64} and applied for solving the VMP problem. It aims to addresses two important Cloud aspects, e.g. efficient energy usage and effective resource utilization.

ii) \emph{GA-based VMP methods}. David Wilcox \textit{et al.} \cite{56} formally defined multi-capacity bin packing, which was the generalization of conventional bin packing. And they also develop an reordering grouping genetic algorithm (RGGA) to assign VMs to servers. A. S. Alashaikh \textit{et al.} \cite{33} adopted the notion of ceteris paribus as an interpretation for the decision maker's preferences and incorporated it in a constrained multi-objective problem known as VMP problems. They proposed a variant of the NSGA-II algorithm that promotes ceteris paribus preferred solutions and evaluate its applicability. X. Xu \textit{et al.} \cite{22} built a many-objective VMP optimization model target to minimize energy consumption and maximize load balance, resource utilization, and robustness. They also proposed an energy-efficient knee point-driven evolutionary algorithm (EEKnEA) to address their optimization model.

iii) \emph{ACS-based VMP methods}. An early study using ACS for VMP problems was introduced by E. Feller \textit{et al.} \cite{60}. In their study, the VMP problem was formulated as a static multiple dimensional bin packing problem, which the optimization goal was to minimize the number of cloud servers for support current load. To solve the formulated optimization problem, an ACS algorithm was also developed in \cite{60} coupled with a max-min updated method. Nevertheless, this method installs VMs in PSs only based on a single resource. Another study adopting ACS for VMP problems was introduced by X. Liu \textit{et al.} \cite{61}, which consolidates VMs according to multiple resources. In their approach, a different approach was designed to update the pheromones between two pairs of VMs to measure their need for PM in ACS. The number of PSs is the same as the number of the VMs in the first generation and it was decreased as the evolutionary proceed. This work on ACS for VMP problem was improved later by X. Liu \textit{et al.} \cite{31}, which involves an proposed order exchange and migration (OEM) local search techniques to transform an infeasible solution into a feasible solution.

In addition to the above three categories, there are other heuristic algorithms for solving VMP problems. For example, N. K. Sharma \textit{et al.} \cite{30} introduced approach of GA and PSO referred to as HGAPSO for the placement policy of VMs to PSs. X. Li \textit{et al.} \cite{62} proposed discrete firefly algorithm to solve VMP problems, which takes firefly's location as the placement result and brightness as the objective value.

\subsection{Multi-Factorial Evolutionary Algorithm (MFEA)}

The multi-factorial evolutionary algorithm (MFEA) is able to evolve two or more tasks simultaneously and accelerate the convergence of each task \cite{45}\cite{53}. The unified representation block is built in MFEA to achieve knowledge transfer among tasks, which include the encoding and decoding operation. The encoding operation is called to build an unified express space denoted as $Y$. The individual $y$ in $Y$ includes the genetic material of all tasks. The dimension of $Y$ can be defined as $D_Y$ = max\{$D_i$\}, where $D_i$ is the dimension of the $i$-th task and $H$ is the number of tasks, $i$ = 1, 2, ..., $H$. Reversely, the decoding step can decipher $y$ into $H$ task-specific solutions with the $i$-th solution $x_i$ being $x_i$ = $y$(1: $D_i$), where $y$(1: $D_i$) retrieves the first $D_i$ dimensions of $y$. Some foundation definition of each individual are described as follow:

$\textbf {\emph{Factorial Cost}}$: The objective function value of individual $p_i$ on task $\tau_i$ is defined as the factorial cost $f_p$ of $p_i$. The factorial costs of $p_i$ on the other tasks are set to infinity.

$\textbf {\emph{Factorial Rank}}$: The factorial rank $r{_i^k}$ of individual $p_i$ is defined as the rank of $p_i$ on the $k$-th task considering all individuals in $P$.

$\textbf {\emph{Skill Factor}}$: The skill factor $\tau_i$ of individual $p_i$ indicates the task on which $p_i$ performs the most effectively, i.e., $\tau_i = {\rm argmin}_k\{r^k_i\}$, where \emph{k} = 1, 2, ..., \emph{K}. For the sake of simplicity, we say an individual $p_i$ is specific to a task if the task is the skill factor of $p_i$.

$\textbf {\emph{Scalar Fitness}}$: The scalar fitness $\varphi_i$ of individual $p_i$ is denoted as the reciprocal of the corresponding factorial rank on task $\tau_i$, i.e., $\varphi_i = 1/r^{\tau_i}_i$. A greater scalar fitness value $\varphi_i$ means that $p_i$ can survive to the next generation at a higher probability.

\begin{algorithm}[h]
  \caption{ The Basic Framework of MFEA}
  \label{a1}
  \begin{algorithmic}[1]
    \Require
     population size, \emph{S}; number of tasks, \emph{K}.
    \Ensure
      a series of solution.
    \State Initialize the population \emph{P}
    \State Evaluate \emph{P}
    \While {\emph{the maximum number of evaluations is not reached}}
        \State Reproduce offspring \emph{O} by assortative matting
        \State Assign the skill factor for \emph{O} by vertical cultural transmission
        \State Evaluate \emph{O}
        \State Generate new population \emph{NP} = \emph{P} $\cup$ \emph{O}
        \State Update the scalar fitness $\varphi$ of \emph{NP}
        \State Select the $P$ fittest individuals from \emph{NP}
    \EndWhile
  \end{algorithmic}
\end{algorithm}

Further, the pseudo code of MFEA is presented in Algorithm \ref{a1}, which can be summarized as follow.

In the beginning, the initial operations, e.g., randomly generating \emph{N} individuals in a unified express space \emph{Y} and assigning the skill factor to each individual in the initial population, are called to form a population. Then, the crossover \cite{54} and polynomial mutation \cite{55} operations are employed to reproduce offspring according to the assortative mating and vertical cultural transmission mechanism. Notably, the assortative mating and vertical cultural transmission are the characteristic and essential components of MFEA, which allow individuals from different tasks to share genetic information with each other. Specifically, the assortative mating mechanism enables the individuals from different tasks to mate with each other at a certain probability, namely random matting probability (\emph{rmp}). The vertical cultural transmission mechanism randomly assigns skill factors to the offspring. It means that the offspring specific to one task may be switched to another directly, leading to a complex optimization task that may acquire superior solutions by learning from other tasks. More details of assortative mating and vertical cultural transmission are available in \cite{45}. Afterward, the factorial cost, factorial rank, and scalar fitness of each offspring individual are updated. Finally, the elite-based environment selection operator is employed to form the next generation population.

In recent years, MFEA and its variants have been successfully applied to various real-world optimization problems. In particular, L. Zhou \textit{et al.} \cite{48} proposed the MFEA equipped with a permutation-based unified representation and split-based decoding operator to solve the NP-hard capacitated vehicle routing problems. H. ThiThanh Binh \textit{et al.} \cite{49} proposed a modified MFEA for cluster shortest path tree problems (CSTP), together with novel genetic operators, e.g., population initialization, crossover, and mutation operators. Furthermore, a novel decoding scheme for deriving factorial solutions from the unified representation in MFEA, which is the critical factor to the performance of any variant of the MFEA, is also introduced in \cite{49}. C. Yang \textit{et al.} \cite{50} extend their method TMO-MFEA proposed in \cite{66} to solve operational indices optimization, which involves a formulated multi-objective multi-factorial operational indices optimization problem. In their proposed optimization model, the most accurate task is considered to be the original functions of the solved problem, while the remained models are the helper tasks to accelerate the optimization of the most accurate task.

\section{The formulation of VMP Problem}

The VMP problems can be seen as linear programming (LP) problems, which usually consider hardware resource constraints of PSs such as CPU, RAM and disk. As the industry's demand for VMs continues to grow, the single-task EA-based solvers have suffered from some limitations when addressing a LVMP problem. Most of them require more optimization time to complete the allocation of VMs onto PSs in the large-scale data center, which resulted in the poor scalability. In this paper, the LVMP problem is reformulated in the form of the MFO problem, which is described as follows. The used variables and their definitions are summarized in Table \ref{t1}.

\begin{equation}
\begin{array}{l}{\operatorname{Min} f(s)=\min T_{1}\left(s_{1}\right) \cup \min T_{2}\left(s_{2}\right) \cup \ldots \cup \min T_{H}\left(s_{H}\right)} \\ \\ {\text { subject to } \quad s=\left\{s_{1}, s_{2}, \ldots, s_{H}\right\}}\end{array}
\end{equation}
where $T_i$ represents a SVMP optimization task that decomposed from the LVMP problem, which is described as follow. And $s_i$ is the solution of $i$-th optimization task $T_i$, $i$ = 1, 2，...，$H$. $H$ is the total number of SVMP optimization tasks.

The main differences between the established MFO problem and other MFO problems lie in two aspects presented as follows. On the one hand, the existing MFO problems aim to solve multiple optimization tasks, and the obtained solution set are corresponding to all tasks. Conversely, the purpose of established MFO problems is to address the single optimization problems. On the other hand, the number of optimization tasks $H$ is changed dynamically. It means that $H$ varies with the size of the VMs in the data center.

Assume that there are $V$ VMs and $M$ PSs in the data center. Given $N_i$ VMs assigned to $i$-th VMP optimization task, the number of VMP optimization tasks $H$ is defined as follows:

\begin{equation}
H=\left\lfloor V / N_{i}\right\rfloor
\end{equation}

Considering that the relationship between $V$ and $N_i$ is not always divisible, the number of VMs assigned to the $H$-th VMP optimization task $N_H$, e.g., $i$ = $H$, is reallocated as follows:

\begin{equation}
N_{H}=\lfloor V / H\rfloor+ V \% H
\end{equation}
where the symbol $\%$ is the mod operation. Owing to the organic composition of different types of PSs, the data center could be operated with high utilization of PS clusters and achieved energy efficiency. However, the existing VMP researches generally designed the optimization model to lower the number of activated PSs or improve the utilization of PSs cluster, which ignore the heterogeneity between PSs. In the actual cloud environment, when a PS is activated, it brings deployment costs that depends on its configuration. It means that the deployment costs can be naturally used to guide which type of PS to activate for improving the utilization of PSs cluster. According to the description above, this paper proposes a deployment cost based optimization model to solve VMP problems in heterogeneous environment.

Assume that $M$ PSs are categorized in to $L$ different types according to the specific configuration, where $L$ $\textgreater$ 1. The number of each type of PSs is denoted as $PS_l$, where $M=\sum_{l=1}^{L} PS_{l}$. Noted that $M_i$ PSs is also assigned to the $i$-th task, which $M_i$ could be given as:

\begin{equation}
M_{i}=\left\{\begin{array}{l}{\sum_{l=1}^{L}\left\lfloor P S_{l} / H\right\rfloor, \text { if } i=1,2, \ldots, H-1} \\ \\ {\sum_{l=1}^{L}\left(\left\lfloor P S_{l} / H\right\rfloor+ P S_{l} \% H\right), \text { if } i=H}\end{array}\right.
\end{equation}

The assignment tag of VMs in $i$-th task is denoted as \textbf{$X_i$}, which is a $N_i \times $\textbf{$M$} matrix consisted of 0/1 variables. If the $VM_j$ in $i$-th task is placed on $PS_k$, then the element $v_{i,j,k}$ of $X_i$  =1, otherwise $v_{i,j,k}$ =0. Similarly, the activated tag of PSs in $i$-th task is denoted as $P_i$ , which is a $\parallel M_i \parallel$ vector consisted of 0/1 variables. If there are greater than or equal to one VM placed in $PS_k$, then the element $p_{i,k}$ of $P_i$ = 1, otherwise $p_{i,k}$ = 0. The element $v_{i,j,k}$ and $p_{i,k}$p are formulated as follows:

\begin{equation}
v_{i, j, k}=\left\{\begin{array}{ll}{1,} & {\text { if } v m_{i, j} \text { is placemented in } p_{i, k}} \\ \\ {0,} & {\text { otherwise }}\end{array}\right.
\end{equation}

\begin{equation}
p_{i, k}=\left\{\begin{array}{ll}{1,} & {\text { if } \sum_{j=1}^{N_{i}} v_{i, j, k} \geq 1} \\ \\ {0,} & {\text { otherwise }}\end{array}\right.
\end{equation}

In this work, three typical computing resources (CPU, RAM and disk) in cloud computing are considered as constraint options. Based on the above description, the deployment cost based SVMP optimization task for the established MFO problem can be given as follows:

\begin{equation}
\begin{array}{l}{\operatorname{Min} T_{i}\left(s_{i}\right)=\sum_{k=1}^{M_{i}} p s_{i, k}^{\operatorname{cost}} \times p_{i, k}} \\ \\
{\text { subject to }} \\ \\ {\quad \sum_{k=1}^{M_{i}} v_{i, j, k}=1} \\ \\
\sum_{j=1}^{N_{i}} v m_{i, j}^{C P U} \times v_{i, j, k} \leq p s_{i, k}^{C P U} \times p_{i, k} \\ \\
\sum_{j=1}^{N_{i}} v m_{i, j}^{R A M} \times v_{i, j, k} \leq p s_{i, k}^{R A M} \times p_{i, k} \\ \\
\sum_{j=1}^{N_{i}} v m_{i, j}^{d i s k} \times v_{i, j, k} \leq p s_{i, k}^{d i s k} \times p_{i, k}

\end{array}
\end{equation}
where $vm_{i, j}^{\text {CPU}}$, $vm_{i, j}^{\text {RAM}}$ and $vm_{i, j}^{\text {disk}}$ represent the computing source requirement CPU, RAM and disk of the $j$-th VM in $i$-th optimization task, respectively.

\begin{table}[H]
\captionsetup{font={scriptsize}}
\caption{DEFINITION OF SYMBOLS}
\label{t1}
\centering
\linespread{1.2}
\small
\begin{tabular}{ll}
\toprule[0.8pt]
Symbols&Definition\\
\toprule[0.8pt]
$V$&Number of VMs in data center\\
$M$&Number of PSs in data center\\
$L$&Number of PSs types in data center\\
$H$&Number of VMP optimization task\\
$T_i$&The $i$-th VMP optimization task\\
$s_i$&Solution of $i$-th VMP optimization task\\
$N_i$&Number of VMs assign to each VMP optimization task\\
$M_i$&Number of PSs assign to $i$-th VMP optimization task\\
$PS_{l}$&Number of $l$-th type PS in data center\\
$\textbf{v}_{i,j,k}$&Binary variable of VM placement\\
$vm_{i,j}$&The $j$-th VM in $i$-th VMP optimization task\\
$vm_{i,j}^{cpu}$&CPU requirement of VM $vm_{i,j}$\\
$vm_{i,j}^{ram}$&RAM requirement of VM $vm_{i,j}$\\
$vm_{i,j}^{disk}$&Disk requirement of VM $vm_{i,j}$\\
$\textbf{p}_{i,k}$&Binary variable of PS status\\
$ps_{i,k}$&The $k$-th PS in data center\\
$ps_{i,k}^{cost}$&Deployment cost of the $k$-th PS\\
$ps_{i,k}^{cpu}$&CPU capacity of PS $ps_k$\\
$ps_{i,k}^{ram}$&RAM capacity of PS $ps_k$\\
$ps_{i,k}^{disk}$&Disk capacity of PS $ps_k$\\
\toprule[0.8pt]
\end{tabular}
\end{table}

\section{Our Proposed Method}

\subsection{The Overall Framework}
The overall framework of our proposed method is described in Algorithm \ref{a2}.

Line 1 to 2 represent the pretreatment process of the input VMs list. The input VMs list is randomly assigned to \emph{H} optimization tasks. Line 3 is the process of building unified express space for \emph{H} VMP optimization tasks, which is presented in subsection IV.B in detail. Lines 4 to 6 describe a series of initialization operator for generating the population in unified express space. Lines 8 to 9 are the process of offspring reproduction, which is similar to the basic MFEA. The crossover and mutation operators used in assortative matting are described detailedly in subsection IV.D. After that, the generated offspring population is assigned the skill factor according to the vertical cultural transmission. Line 10 is the proposed greedy-based allocation operator, which is described in subsection IV.C. Line 11 is the evaluation of offspring population and lines 12 to 13 represent the updated the scalar fitness for both the offspring and parent population. Line 14 forms the population to survive into the next generation based on the elite-based environment selection. Lines 7 to 15 are executed repeatedly until the termination condition is satisfied. Line 16 is the proposed re-migration and merge operator, which is called to obtained the placement solution of the input VMs list.

\begin{algorithm}[h]
  \caption{The Framework of Our proposed Method}
  \label{a2}
  \begin{algorithmic}[1]
    \Require
     VMs List, \emph{V};
     PSs List, \emph{M};
     Number of PSs types, \emph{L};
     Number of VMs assigned to each task, \emph{N}.
    \Ensure
     The placement solution for \emph{V}.
    \State Obtain the number of tasks \emph{H} according to Eq. (2)
    \State Assign \emph{V} to \emph{H} tasks randomly
    \State Build unified express space among \emph{H} tasks
    \State Initialize the population \emph{P}
    \State Perform greedy-based allocation operator on \emph{P} $\leftarrow$ Algorithm 3.
    \State Evaluate \emph{P}
    \While {\emph{the maximum number of evaluations is not reached}}
        \State Reproduce offspring \emph{O} by assortative matting
        \State Assign the skill factor for \emph{O} by vertical cultural transmission
        \State Perform greedy-based allocation operator on \emph{O} $\leftarrow$ Algorithm 3.
        \State Evaluate \emph{O}
        \State Generate new population \emph{NP} = \emph{P} $\cup$ \emph{O}
        \State Update the scalar fitness $\varphi$ of \emph{NP}
        \State Select the $P$ fittest individuals from \emph{NP}
    \EndWhile
    \State Perform re-migration and merge operator on \emph{P} $\leftarrow$ Algorithm 4.
  \end{algorithmic}
\end{algorithm}

\subsection{Building Unified Express Space}
In this subsection, the procedure of building the unified express space is described as follows.

To achieve the implicit knowledge transfer among \emph{H} VMP optimization tasks, it is essential to construct a unified genetic express space among them. The individuals, which are generated in unified express space, contain all the information of VMs for each task. Note that the number of the same VM types is not equal in each VMP optimization task due to the random assignment process. Different types of VM refer to the different configuration (such as CPU, RAM, or disk). For each different types of VMs, take the maximum number of VMs in all tasks as the number of this types of VMs in unified express space.

\begin{figure}
\captionsetup{font={footnotesize}}
\centering
\includegraphics[scale=0.35]{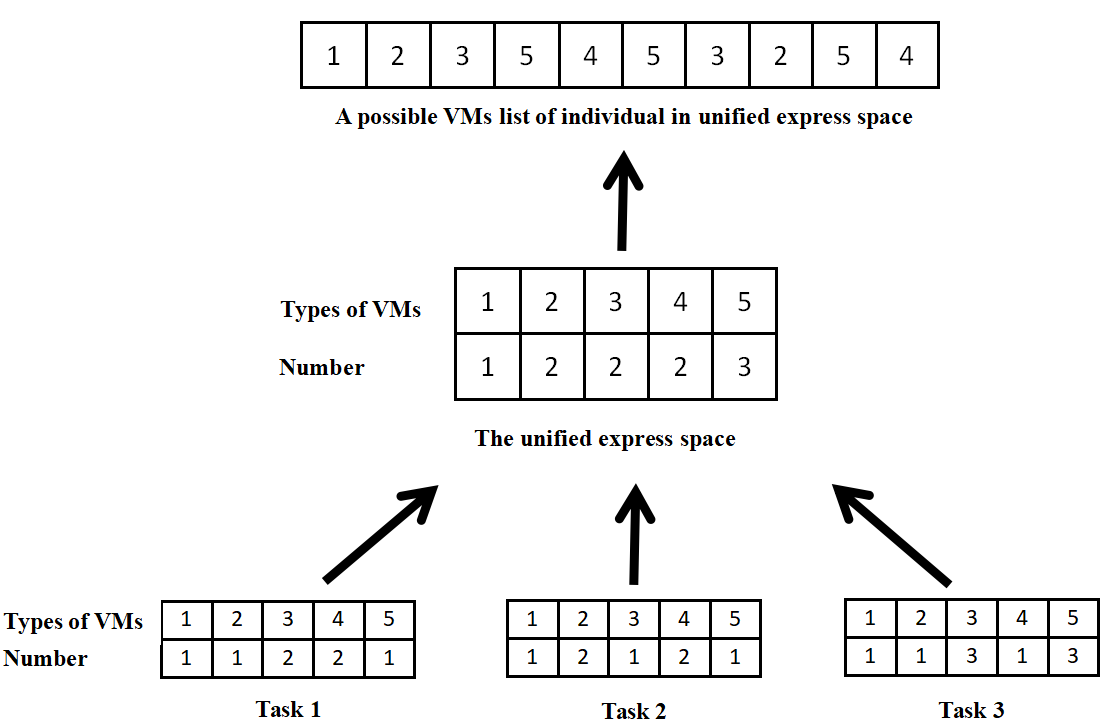}
\captionsetup{justification=centering}
\caption{The process of building unified express space}
\label{f3}
\end{figure}

Fig. \ref{f3} illustrates an example of establishing the unified express space. Suppose that there are five types of VMs in the data center. Their type ID is denoted as \emph{type-1}, \emph{type-2},..., \emph{type-5}, respectively. And all the VMs are assigned to three VMP optimization tasks randomly. Take the \emph{type-2} as an example, the number of \emph{type-2} assigned to the first VMP optimization task is 1, the number of \emph{type-2} assigned to the second VMP optimization task is 2, and the number of \emph{type-2} assigned to the third VMP optimization task is set to be 2. As a result, the number of \emph{type-2} is 2 in unified express space. Other types of VMs do the same operation. After the transform process above, the individuals generated in unified express space retain the information of each VMP optimization task and it could be decoded into all tasks.

\begin{figure}
\captionsetup{font={footnotesize}}
\centering
\includegraphics[scale=0.5]{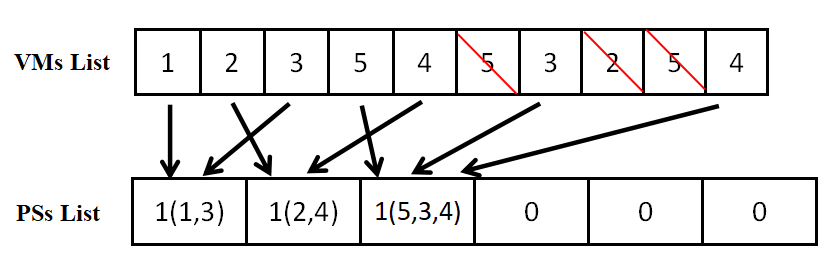}
\captionsetup{justification=centering}
\caption{The representation block of individual in unified express space}
\label{f4}
\end{figure}

The representation block of individual in unified express space is shown in Fig. \ref{f4}. In a individual, if the PS contains VMs, it is assigned 1, otherwise 0. The PSs list not only consists of binary number, but records the information of VMs with which it loaded. In Fig. \ref{f4}, for example, the symbol of first PS is assigned 1, which indicates that the first PS is activated and installs the \emph{type-1} and \emph{type-3} VMs.

Compared with the single-factorial solver, the representation block of individual in our method is more effective. Take Fig. \ref{f3} as an example, suppose there are 23 VMs in the data center. An possible individual in the unified express space is shown in Fig. \ref{f3}, it can be seen that each individual generated in unified express space only needs to contain the information of 10 VMs. And the representation block of individual in single-factorial solver is consisted of the information of 23 VMs. The individual with fewer information of VMs means that it can complete the placement of VMs onto PSs in a faster time obviously.

\subsection{Greedy-based Allocation Operator}
In this subsection, the proposed greedy-based allocation operator is described as follows.

\begin{figure}
\captionsetup{font={footnotesize}}
\centering
\includegraphics[scale=0.4]{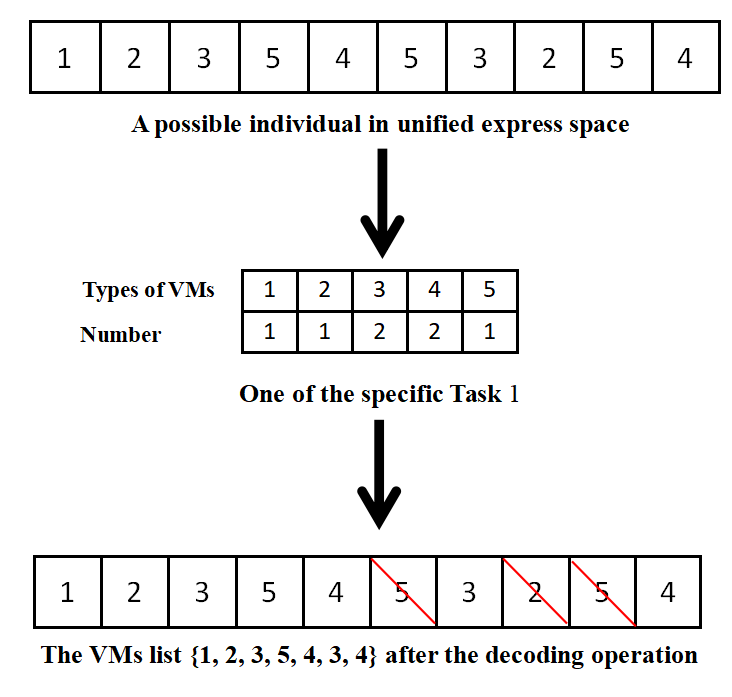}
\captionsetup{justification=centering}
\caption{An example of decoding operation}
\label{f11}
\end{figure}

The VMs list of individual in the unified express space need to be decoded into its corresponding task before the greedy-based allocation operator is performed. The decoding operation takes the first \emph{n} VMs that meet the types and number requirement of VMs assigned to a task to generate the VMs list. Then the greedy-based allocation operator is performed on the generated VMs list. An example of the decoding operation is depicted in Fig. \ref{f11}. As shown in Fig. \ref{f11}, a possible VMs list of individual in unified express space is \{1, 2, 3, 5, 4, 5, 3, 2, 5, 4\}. After the decoding operation, the generated VMs list for one of the specific task is \{1, 2, 3, 5, 4, 3, 4\}.

The pseudo code of greedy-based allocation operator is summarized in Algorithm \ref{a4}. The data center has a variety of different configuration PSs. For each type PS, we select one to load the generated VMs list until any specific resource (CPU, RAM, or disk) of the selected PS is exhausted or there are no VMs can be loaded in the selected PS. Then we select the PS with the highest comprehensive utilization rate and erase the VMs it loaded from the generated VMs list. The definition of comprehensive utilization rate \emph{U} of a single PS is below:

\begin{equation}
\begin{array}{l}{U = a \times U_{C P U}+b \times U_{R A M}+c \times U_{d i s k}} \\ \\ {\text { subject to } \quad a+b+c=1}\end{array}
\end{equation}
where the parameters $a$, $b$, and $c$ are the weight coefficient. To ensure the importance of three selected hardware resource is the same, we set $a$, $b$ and $c$ as the same weight to guarantee the fairness of them, i.e., $a$ = $b$ = $c$ = 1/3. The utilization of CPU $U_{CPU}(x)$, the utilization of RAM $U_{RAM}(x)$ and the utilization of disk $U_{disk}(x)$ are given as follows:

\begin{equation}
U_{\text {CPU}}=\frac{\sum_{j=1}^{N_{i}} v m_{i, j}^{\text {CPU}} \times v_{i, j, k}}{ps_{i, k}^{\text {CPU}}}, \forall i \in H \wedge \forall j \in N_{i} \wedge \forall k \in M_{i}
\end{equation}

\begin{equation}
U_{\text {RAM}}=\frac{\sum_{j=1}^{N_{i}} v m_{i, j}^{\text {RAM}} \times v_{i, j, k}}{ps_{i, k}^{\text {RAM}}}, \forall i \in H \wedge \forall j \in N_{i} \wedge \forall k \in M_{i}
\end{equation}

\begin{equation}
U_{\text {disk}}=\frac{\sum_{j=1}^{N_{i}} v m_{i, j}^{\text {disk}} \times v_{i, j, k}}{ps_{i, k}^{\text {disk}}}, \forall i \in H \wedge \forall j \in N_{i} \wedge \forall k \in M_{i}
\end{equation}
where $ps_{i, k}^{CPU}$, $ps_{i, k}^{RAM}$ and $ps_{i, k}^{disk}$ are the CPU, RAM and disk resource capacity of $k$-th PS, respectively.

\begin{algorithm}[h]
  \caption{Greedy-based Allocation Operator}
  \label{a4}
  \begin{algorithmic}[1]
    \Require
     VMs List of a individual, \emph{V};
     PSs List of a individual, \emph{M};
     Number of PSs types, \emph{L}.
    \Ensure
     The placement solution for \emph{V}.
    \While {\emph{V is not empty}}
        \State \emph{psList} = []
        \For{i = 1 to \emph{L}}
            \If {the remaining amount of \emph{M}[i] \textgreater 0}
              \For j = 1 to \emph{V}
                \If {$V$[j] can be placed in \emph{ps}}
                    \State Add $V$[j] to \emph{M}[i]
                \EndIf
                \If {\emph{M}[i] is running out of resources}
                    \State break;
                \EndIf
              \EndFor
              \State Calculate the utilization of \emph{M}[i]
              \State Add \emph{M}[i] into $psList$
            \EndIf
        \EndFor
        \State Select the \emph{ps} with the highest utilization in \emph{psList}
        \State Erase VMs placed in \emph{ps} from \emph{V}
        \State \emph{M}[ps] = \emph{M}[ps] - 1
    \EndWhile
  \end{algorithmic}
\end{algorithm}

\subsection{Crossover and Mutation Operator}

The offspring population is reproduced by assortative mating. The crossover and mutation operator used in assortative mating are described in detail as follow.

Crossover operator: The crossover operator aims to reproduce offspring population theoretically with good traits from both parents and hopefully better fitness \cite{56}. In this paper, the crossover operator is implemented in an modified exon shuffling approach similar to the one described by \cite{57}. This approach combines all the activated PSs from randomly selected parents and sorts the PSs by comprehensive utilization, which is calculated based on Eq. (8). The more full PSs are at the front of the list, while the less full PSs are at the end. The crossover operator systematically picks the more full PSs and keeps the VMs that have loaded intact. If the picked PS contains any VMs that have been installed in other PS, the picked PS is discarded and set to be idle status. The remaining VMs that have not been placed in any PSs are randomly disrupted and inserted to the end of the generated VM list. This crossover approach preserves as many VM lists as possible in more full-filled PSs. It also ensures that the generated candidate solutions are all feasible solutions and it is avail to speed up the running time of the algorithm.

Mutation operator: After the crossover operator, each individual is allowed to having their candidate solution modified slightly at a certain probability. The mutation operator aims to facilitate the population to escape from local optima. In this paper, it is realized by randomly swapping the position of two VMs in the VMs list of individual.

\subsection{Fitness Evaluation}
After the individual has finished the placement process, we evaluate it fitness according to the status of PSs (activated or idle). The fitness evaluation function is define as the deployment cost of activated PSs list of individual in $i$-th task, which is given as follows:

\begin{equation}
f\left(s_ip\right)=\sum_{x=1}^{L} \sum_{j=1}^{M_{i}} \cos t_{x} \times p_{i, j}^{x}
\end{equation}
where $cost_x$ is the deployment cost of $x$-th type PS. After the fitness evaluation process, the \emph{scalar fitness} of the individual in population that consists of the parent and offspring is updated. Then, in the elite-based environment selection operator, the individuals with smaller \emph{scalar fitness} are survived to the next generation. In particular, the number of individuals from different tasks survived to the next generation is equal.

\subsection{Re-migration and Merge operator}
Although the obtained solutions of each tasks is feasible, they are not the solutions of the original input VMs list. The designed re-migration and merge operator is called to offer the solution of the original input LVMP problems from the obtained solutions of MFO problem, which the pseudo code is presented in Algorithm \ref{a5}.

For the fittest solution in each VMP optimization task, the re-migration operator is employed to the PS list that are not fully utilized by computing resources, e.g., the available computing resources are left behind. Specifically, the VMs that loaded in the PS which have the remaining compute resource are re-popped, which the new VMs list is generated to be refilled by the greedy-based allocation operator. Finally, the merge operator is executed to combine the activated PSs among all the fittest solution and form the placement scheme of the LVMP problem.

\begin{algorithm}[h]
  \caption{Re-migration and Merge Operator}
  \label{a5}
  \begin{algorithmic}[1]
    \Require
     Best individual of each task, $C$; Number of tasks, $H$.
    \Ensure
     The placement solution $S$.
    \State Solution $S$ = []
    \State $vmList$ = []
    \For{i = 1 to $H$}
        \For{j = 1 to $ps$ in $C[i]$ }
        \If {every resource has a surplus in $ps_j$}
            \State Remove $vms$ from $ps_j$
            \State Insert $vms$ into $vmList$
        \Else
            \State Add $ps_j$ to $S$
        \EndIf
    \EndFor
    \EndFor

    \State Backfill $vmList$ to $S$ $\leftarrow$ Algorithm 3.
  \end{algorithmic}
\end{algorithm}

\subsection{Time Complexity Analysis}
Besides the capability of finding solutions of high quality, time complexity is a significant issue for an optimization algorithm. The time complexity of our proposed method is mainly based on GA. Assume the maximum iteration is ``\emph{I}", the number of PSs is ``\emph{M}", the number of PSs types is ``\emph{L}", the number of VMs is ``\emph{V}", and the generation size is ``\emph{G}". The time complexity of our method is mainly depended on the offspring reproduction and greedy-based allocation operator. Further, as seen in Algorithm \ref{a4}, the time complexity of proposed greedy-based allocation operator is $O$(iteration $\ast$ generation size $\ast$ the number of VMs $\ast$ the number of PSs types $\ast$ the number of VMs), which is $O$(\emph{I} $\ast$ \emph{G} $\ast$ \emph{V} $\ast$ \emph{L} $\ast$ \emph{V}). And the the offspring reproduction require $O$(\emph{I} $\ast$ \emph{G} $\ast$ \emph{M} $\ast$ \emph{V}) time complexity. Since the number of VMs is larger than the number of PSs, based on the analysis above, the time complexity of our method is $O$(\emph{I} $\ast$ \emph{G} $\ast$ \emph{V} $\ast$ \emph{V} $\ast$ \emph{L}).

\section{Experiments}	

Simulation Experimental tests are carried out in this section to assess the performance of our proposed. All the compared algorithms have been implemented in C++, and ran on a PC with a Pentium Dual CPU i7 and 8.0 GB RAM.

\subsection{Test Data Set Introduction}	

\renewcommand{\arraystretch}{0.5} 
\linespread{1.2}
\begin{table}[tp]
  \centering
  \fontsize{8}{10}\selectfont
  \begin{threeparttable}
  \captionsetup{font={footnotesize}}
  \caption{The configuration of Physical Servers}
  \label{t2}
      \begin{tabular}{lcccc}
      \toprule
      Type&CPU&RAM&Disk&Cost\\
      \midrule
      General&\makecell[tl]{56-core}&\makecell[tl]{128G}&\makecell[tl]{1200G}&\makecell[tl]{3.49}\cr
      LargeRAM&\makecell[tl]{84-core}&\makecell[tl]{256G}&\makecell[tl]{2400G}&\makecell[tl]{4.36}\cr
      HighPerformance&\makecell[tl]{112-core}&\makecell[tl]{192G}&\makecell[tl]{3600G}&\makecell[tl]{5.45}\cr
      \bottomrule
      \end{tabular}
  \end{threeparttable}
\end{table}

In this work, the resources characterize of VMs and PSs are downloaded from the Huawei cloud website\footnote{https://activity.huaweicloud.com/promotion/index.html}. Three different types of PSs are adopted to build heterogeneous environments, which are respectively denoted as \emph{General}, \emph{LargeRAM} and \emph{HightPerformance}. The configurations of PSs are shown in Table \ref{t2}. Further, these three different types of PSs have different bias characteristics. \emph{LargeRAM} has a larger memory, preferring to place the VM with greater RAM demand. \emph{HightPerformance} has more CPU cores and is suitable for the VMs that required high computing power. The resources characterized of \emph{General} is between the above two. Besides, we use 100 different types of VMs for the creation of large-scale test data sets, and the configuration of VMs are shown in Table \ref{t10}. The ratio of CPU and RAM of each VM comes from the Huawei cloud website. The configuration of disk is set to 100-500 (G) and it is randomly generated at 100 (G) intervals, which is considered as bottleneck resource in this experiment. The total ratio of the requirement of CPU, RAM, and disk is approximately 6:9:10 on each test data set, which proves the resource of disk is served as a bottleneck resource in this experiment. In conclusion, taking the data set with the size of 5,000 VMs as an example, it is composed of 5,000 VMs which randomly produced from 100 different types of VMs with the discrete uniform distribution. Point out that the optimal solution to the test data sets are not clear due to the random creation process, which could be served as black-box testing problems.

\renewcommand{\arraystretch}{0.5} 
\linespread{1.2}
\begin{table*}[tp]
  \centering
  \small
  \fontsize{10}{9.8}\selectfont
  \begin{threeparttable}
  \captionsetup{font={normal}}
  \caption{CONFIGURATION OF VIRTUAL MACHINES}
  \label{t10}
    \begin{tabular}{cccccccccccc}
    \toprule
    Type&CPU&RAM&disk&Type&CPU&RAM&disk&Type&CPU&RAM&disk\\
    &(core)&(G)&(G)&&(core)&(G)&(G)&&(core)&(G)&(G)\\
    \midrule
    \makecell[tl]{\emph{type-1}}&\makecell[tl]{1}&\makecell[tl]{1}&\makecell[tl]{100}&\makecell[tl]{\emph{type-36}}&\makecell[tl]{2}&\makecell[tl]{16}&\makecell[tl]{100}&\makecell[tl]{\emph{type-71}}&\makecell[tl]{8}&\makecell[tl]{32}&\makecell[tl]{100}\cr
    \makecell[tl]{\emph{type-2}}&\makecell[tl]{1}&\makecell[tl]{1}&\makecell[tl]{200}&\makecell[tl]{\emph{type-37}}&\makecell[tl]{2}&\makecell[tl]{16}&\makecell[tl]{200}&\makecell[tl]{\emph{type-72}}&\makecell[tl]{8}&\makecell[tl]{32}&\makecell[tl]{200}\cr
    \makecell[tl]{\emph{type-3}}&\makecell[tl]{1}&\makecell[tl]{1}&\makecell[tl]{300}&\makecell[tl]{\emph{type-38}}&\makecell[tl]{2}&\makecell[tl]{16}&\makecell[tl]{300}&\makecell[tl]{\emph{type-73}}&\makecell[tl]{8}&\makecell[tl]{32}&\makecell[tl]{300}\cr
    \makecell[tl]{\emph{type-4}}&\makecell[tl]{1}&\makecell[tl]{1}&\makecell[tl]{400}&\makecell[tl]{\emph{type-39}}&\makecell[tl]{2}&\makecell[tl]{16}&\makecell[tl]{400}&\makecell[tl]{\emph{type-74}}&\makecell[tl]{8}&\makecell[tl]{32}&\makecell[tl]{400}\cr
    \makecell[tl]{\emph{type-5}}&\makecell[tl]{1}&\makecell[tl]{1}&\makecell[tl]{500}&\makecell[tl]{\emph{type-40}}&\makecell[tl]{2}&\makecell[tl]{16}&\makecell[tl]{500}&\makecell[tl]{\emph{type-75}}&\makecell[tl]{8}&\makecell[tl]{32}&\makecell[tl]{500}\cr
    \makecell[tl]{\emph{type-6}}&\makecell[tl]{1}&\makecell[tl]{2}&\makecell[tl]{100}&\makecell[tl]{\emph{type-41}}&\makecell[tl]{4}&\makecell[tl]{4}&\makecell[tl]{100}&\makecell[tl]{\emph{type-76}}&\makecell[tl]{8}&\makecell[tl]{64}&\makecell[tl]{100}\cr
    \makecell[tl]{\emph{type-7}}&\makecell[tl]{1}&\makecell[tl]{2}&\makecell[tl]{200}&\makecell[tl]{\emph{type-42}}&\makecell[tl]{4}&\makecell[tl]{4}&\makecell[tl]{200}&\makecell[tl]{\emph{type-77}}&\makecell[tl]{8}&\makecell[tl]{64}&\makecell[tl]{200}\cr
    \makecell[tl]{\emph{type-8}}&\makecell[tl]{1}&\makecell[tl]{2}&\makecell[tl]{300}&\makecell[tl]{\emph{type-43}}&\makecell[tl]{4}&\makecell[tl]{4}&\makecell[tl]{300}&\makecell[tl]{\emph{type-78}}&\makecell[tl]{8}&\makecell[tl]{64}&\makecell[tl]{300}\cr
    \makecell[tl]{\emph{type-9}}&\makecell[tl]{1}&\makecell[tl]{2}&\makecell[tl]{400}&\makecell[tl]{\emph{type-44}}&\makecell[tl]{4}&\makecell[tl]{4}&\makecell[tl]{400}&\makecell[tl]{\emph{type-79}}&\makecell[tl]{8}&\makecell[tl]{64}&\makecell[tl]{400}\cr
    \makecell[tl]{\emph{type-10}}&\makecell[tl]{1}&\makecell[tl]{2}&\makecell[tl]{500}&\makecell[tl]{\emph{type-45}}&\makecell[tl]{4}&\makecell[tl]{4}&\makecell[tl]{500}&\makecell[tl]{\emph{type-80}}&\makecell[tl]{8}&\makecell[tl]{64}&\makecell[tl]{500}\cr
    \makecell[tl]{\emph{type-11}}&\makecell[tl]{1}&\makecell[tl]{4}&\makecell[tl]{100}&\makecell[tl]{\emph{type-46}}&\makecell[tl]{4}&\makecell[tl]{8}&\makecell[tl]{100}&\makecell[tl]{\emph{type-81}}&\makecell[tl]{16}&\makecell[tl]{16}&\makecell[tl]{100}\cr
    \makecell[tl]{\emph{type-12}}&\makecell[tl]{1}&\makecell[tl]{4}&\makecell[tl]{200}&\makecell[tl]{\emph{type-47}}&\makecell[tl]{4}&\makecell[tl]{8}&\makecell[tl]{200}&\makecell[tl]{\emph{type-82}}&\makecell[tl]{16}&\makecell[tl]{16}&\makecell[tl]{200}\cr
    \makecell[tl]{\emph{type-13}}&\makecell[tl]{1}&\makecell[tl]{4}&\makecell[tl]{300}&\makecell[tl]{\emph{type-48}}&\makecell[tl]{4}&\makecell[tl]{8}&\makecell[tl]{300}&\makecell[tl]{\emph{type-83}}&\makecell[tl]{16}&\makecell[tl]{16}&\makecell[tl]{300}\cr
    \makecell[tl]{\emph{type-14}}&\makecell[tl]{1}&\makecell[tl]{4}&\makecell[tl]{400}&\makecell[tl]{\emph{type-49}}&\makecell[tl]{4}&\makecell[tl]{8}&\makecell[tl]{400}&\makecell[tl]{\emph{type-84}}&\makecell[tl]{16}&\makecell[tl]{16}&\makecell[tl]{400}\cr
    \makecell[tl]{\emph{type-15}}&\makecell[tl]{1}&\makecell[tl]{4}&\makecell[tl]{500}&\makecell[tl]{\emph{type-50}}&\makecell[tl]{4}&\makecell[tl]{8}&\makecell[tl]{500}&\makecell[tl]{\emph{type-85}}&\makecell[tl]{16}&\makecell[tl]{16}&\makecell[tl]{500}\cr
    \makecell[tl]{\emph{type-16}}&\makecell[tl]{1}&\makecell[tl]{8}&\makecell[tl]{100}&\makecell[tl]{\emph{type-51}}&\makecell[tl]{4}&\makecell[tl]{16}&\makecell[tl]{100}&\makecell[tl]{\emph{type-86}}&\makecell[tl]{16}&\makecell[tl]{32}&\makecell[tl]{100}\cr
    \makecell[tl]{\emph{type-17}}&\makecell[tl]{1}&\makecell[tl]{8}&\makecell[tl]{200}&\makecell[tl]{\emph{type-52}}&\makecell[tl]{4}&\makecell[tl]{16}&\makecell[tl]{200}&\makecell[tl]{\emph{type-87}}&\makecell[tl]{16}&\makecell[tl]{32}&\makecell[tl]{200}\cr
    \makecell[tl]{\emph{type-18}}&\makecell[tl]{1}&\makecell[tl]{8}&\makecell[tl]{300}&\makecell[tl]{\emph{type-53}}&\makecell[tl]{4}&\makecell[tl]{16}&\makecell[tl]{300}&\makecell[tl]{\emph{type-88}}&\makecell[tl]{16}&\makecell[tl]{32}&\makecell[tl]{300}\cr
    \makecell[tl]{\emph{type-19}}&\makecell[tl]{1}&\makecell[tl]{8}&\makecell[tl]{400}&\makecell[tl]{\emph{type-54}}&\makecell[tl]{4}&\makecell[tl]{16}&\makecell[tl]{400}&\makecell[tl]{\emph{type-89}}&\makecell[tl]{16}&\makecell[tl]{32}&\makecell[tl]{400}\cr
    \makecell[tl]{\emph{type-20}}&\makecell[tl]{1}&\makecell[tl]{8}&\makecell[tl]{500}&\makecell[tl]{\emph{type-55}}&\makecell[tl]{4}&\makecell[tl]{16}&\makecell[tl]{500}&\makecell[tl]{\emph{type-90}}&\makecell[tl]{16}&\makecell[tl]{32}&\makecell[tl]{500}\cr
    \makecell[tl]{\emph{type-21}}&\makecell[tl]{2}&\makecell[tl]{2}&\makecell[tl]{100}&\makecell[tl]{\emph{type-56}}&\makecell[tl]{4}&\makecell[tl]{32}&\makecell[tl]{100}&\makecell[tl]{\emph{type-91}}&\makecell[tl]{16}&\makecell[tl]{64}&\makecell[tl]{100}\cr
    \makecell[tl]{\emph{type-22}}&\makecell[tl]{2}&\makecell[tl]{2}&\makecell[tl]{200}&\makecell[tl]{\emph{type-57}}&\makecell[tl]{4}&\makecell[tl]{32}&\makecell[tl]{200}&\makecell[tl]{\emph{type-92}}&\makecell[tl]{16}&\makecell[tl]{64}&\makecell[tl]{200}\cr
    \makecell[tl]{\emph{type-23}}&\makecell[tl]{2}&\makecell[tl]{2}&\makecell[tl]{300}&\makecell[tl]{\emph{type-58}}&\makecell[tl]{4}&\makecell[tl]{32}&\makecell[tl]{300}&\makecell[tl]{\emph{type-93}}&\makecell[tl]{16}&\makecell[tl]{64}&\makecell[tl]{300}\cr
    \makecell[tl]{\emph{type-24}}&\makecell[tl]{2}&\makecell[tl]{2}&\makecell[tl]{400}&\makecell[tl]{\emph{type-59}}&\makecell[tl]{4}&\makecell[tl]{32}&\makecell[tl]{400}&\makecell[tl]{\emph{type-94}}&\makecell[tl]{16}&\makecell[tl]{64}&\makecell[tl]{400}\cr
    \makecell[tl]{\emph{type-25}}&\makecell[tl]{2}&\makecell[tl]{2}&\makecell[tl]{500}&\makecell[tl]{\emph{type-60}}&\makecell[tl]{4}&\makecell[tl]{32}&\makecell[tl]{500}&\makecell[tl]{\emph{type-95}}&\makecell[tl]{16}&\makecell[tl]{64}&\makecell[tl]{500}\cr
    \makecell[tl]{\emph{type-26}}&\makecell[tl]{2}&\makecell[tl]{4}&\makecell[tl]{100}&\makecell[tl]{\emph{type-61}}&\makecell[tl]{8}&\makecell[tl]{8}&\makecell[tl]{100}&\makecell[tl]{\emph{type-96}}&\makecell[tl]{16}&\makecell[tl]{128}&\makecell[tl]{100}\cr
    \makecell[tl]{\emph{type-27}}&\makecell[tl]{2}&\makecell[tl]{4}&\makecell[tl]{200}&\makecell[tl]{\emph{type-62}}&\makecell[tl]{8}&\makecell[tl]{8}&\makecell[tl]{200}&\makecell[tl]{\emph{type-97}}&\makecell[tl]{16}&\makecell[tl]{128}&\makecell[tl]{200}\cr
    \makecell[tl]{\emph{type-28}}&\makecell[tl]{2}&\makecell[tl]{4}&\makecell[tl]{300}&\makecell[tl]{\emph{type-63}}&\makecell[tl]{8}&\makecell[tl]{8}&\makecell[tl]{300}&\makecell[tl]{\emph{type-98}}&\makecell[tl]{16}&\makecell[tl]{128}&\makecell[tl]{300}\cr
    \makecell[tl]{\emph{type-29}}&\makecell[tl]{2}&\makecell[tl]{4}&\makecell[tl]{400}&\makecell[tl]{\emph{type-64}}&\makecell[tl]{8}&\makecell[tl]{8}&\makecell[tl]{400}&\makecell[tl]{\emph{type-99}}&\makecell[tl]{16}&\makecell[tl]{128}&\makecell[tl]{400}\cr
    \makecell[tl]{\emph{type-30}}&\makecell[tl]{2}&\makecell[tl]{4}&\makecell[tl]{500}&\makecell[tl]{\emph{type-65}}&\makecell[tl]{8}&\makecell[tl]{8}&\makecell[tl]{500}&\makecell[tl]{\emph{type-100}}&\makecell[tl]{16}&\makecell[tl]{128}&\makecell[tl]{500}\cr
    \makecell[tl]{\emph{type-31}}&\makecell[tl]{2}&\makecell[tl]{8}&\makecell[tl]{100}&\makecell[tl]{\emph{type-66}}&\makecell[tl]{8}&\makecell[tl]{16}&\makecell[tl]{100}\cr
    \makecell[tl]{\emph{type-32}}&\makecell[tl]{2}&\makecell[tl]{8}&\makecell[tl]{200}&\makecell[tl]{\emph{type-67}}&\makecell[tl]{8}&\makecell[tl]{16}&\makecell[tl]{200}\cr
    \makecell[tl]{\emph{type-33}}&\makecell[tl]{2}&\makecell[tl]{8}&\makecell[tl]{300}&\makecell[tl]{\emph{type-68}}&\makecell[tl]{8}&\makecell[tl]{16}&\makecell[tl]{300}\cr
    \makecell[tl]{\emph{type-34}}&\makecell[tl]{2}&\makecell[tl]{8}&\makecell[tl]{400}&\makecell[tl]{\emph{type-69}}&\makecell[tl]{8}&\makecell[tl]{16}&\makecell[tl]{400}\cr
    \makecell[tl]{\emph{type-35}}&\makecell[tl]{2}&\makecell[tl]{8}&\makecell[tl]{500}&\makecell[tl]{\emph{type-70}}&\makecell[tl]{8}&\makecell[tl]{16}&\makecell[tl]{500}\cr
    \bottomrule
    \end{tabular}
    \end{threeparttable}
\end{table*}

\subsection{Compared Algorithms}	
In this experiment, the effectiveness of the proposed method is validated in comparison with other four different types of state-of-art heuristic algorithms, namely HGAPSO \cite{30}, OEMACS \cite{31}, RGGA \cite{56}, and first-fit decreasing (FFD) \cite{67}. OEMACS construct the assignment of VMs by artificial ants based on the proposed global pheromone updating approach. At the same time, the proposed order exchange and migration (OEM) technique in OEMACS is able to turn the infeasible solutions into feasible solutions, which aims to improve the quality of the solution. HGAPSO proposed a hybrid method of genetic algorithm (GA) and particle swarm optimization (PSO) to complete the allocation of VMs to PSs. The RGGA used the approach of exon shuffling to produce offspring. It sorts the PSs list by utilization rate in descending order and systematically picks the more full PSs to keep them intact.  Finally, the FFD algorithm arranged the VMs list by first considering CPU requirement, then RAM requirement, and lastly disk requirement in descending order. And it installs the VMs in the first PS with adequate computing resource in the PSs list.

\subsection{Experiment Setup}
In this experiment, the parameters setting of the proposed method are the random mating probability (\emph{rmp}) and the number of VMs assigned to $i$-th VMP optimization task $N_i$, which \emph{rmp} = 0.3 and $N_i$ = 200. The above parameters settings are derived from the sensitivity test experiment, which is presented in subsection V.E. The number of maximal iterations is set to 50, the number of individuals $n$ in per task is set to $n$ = 5, the total population size $N$ is the task number $H$ multiplied by $n$, that is, $N = n \times H$.  In addition, the parameters setting of compared algorithm HGAPSO, OEMACS and RGGA set consistently with their original references, with their population size are 10, 5 and 75, and the maximal iteration are 50, 50 and 100, respectively. To conclude, the maximal fitness evaluation of HGAPSO, OEMACS and RGGA are 500, 250 and 7500, respectively. And the fitness evaluation of our method depends on the number of tasks, which is $50 \times n \times H$. The experimental results of all algorithms are obtained over 30 independent runs on each instance.

Numerous performance indicators are used to validate the effectiveness of the proposed method comprehensively. They are the running time of each compared algorithm, the number of activated PSs, the comprehensive utilization rate of activated PS cluster, and the total deployment cost. Node that the comprehensive utilization rate of activated PS cluster is calculated by the Eq. (8). Besides these, the utilization of CPU, the utilization of RAM and the utilization of disk are also adopted as performance indicators in this experiment, which is calculated by the Eq. (9), (10) and (11), respectively.

\subsection{Results and Analysis}

\renewcommand{\arraystretch}{0.7} 
\linespread{1.5}
\begin{table*}[tp]
  \setlength{\tabcolsep}{1mm}{
  \fontsize{5.5}{9}\selectfont
  \begin{threeparttable}
  \captionsetup{font={small}}
  \caption{EXPERIMENTAL RESULT COMPARISONS WITH DIFFERENT SIZE OF VMs}
  \label{t3}
    \begin{tabular}{ccccccccccccccccccccc}
    \toprule
    \multirow{1}{*}{Data Set NO.}&\multirow{1}{*}{Size}&
    \multicolumn{4}{c}{Our Method}&\multicolumn{4}{c}{HGAPSO}&\multicolumn{4}{c}{OEMACS}&\multicolumn{4}{c}{RGGA}&\multicolumn{3}{c}{FFD}\cr
    \cmidrule(lr){3-6} \cmidrule(lr){7-10} \cmidrule(lr){11-14}\cmidrule(lr){15-18}\cmidrule(lr){19-21}
    &&Time&Util&Num&Cost&Time&Util&Num&Cost&Time&Util&Num&Cost&Time&Util&Num&Cost&Util&Num&Cost\cr
    \midrule
    \renewcommand{\arraystretch}{2}
    DS1&5000&\textbf{\makecell[c]{0.74s}}&\textbf{\makecell[c]{85.47\%}}&\textbf{\makecell[c]{534.26}}&\textbf{\makecell[c]{2316.15}}&\makecell[c]{3.29s}&\makecell[c]{75.27\%}&\makecell[tl]{676.70}&\makecell[c]{3079.51}&\makecell[c]{155.02s}&\makecell[c]{79.58\%}&\makecell[c]{643.19}&\makecell[c]{2915.69}&\makecell[c]{29.28s}&\makecell[c]{81.28\%}&\makecell[c]{613.60}&\makecell[c]{2593.51}&\makecell[c]{58.31\%}&\makecell[c]{1081.13}&\makecell[c]{4224.34}\cr\hline
    DS2&10000&\textbf{\makecell[c]{1.48s}}&\textbf{\makecell[c]{85.19\%}}&\textbf{\makecell[c]{1056.01}}&\textbf{\makecell[c]{4578.87}}&\makecell[c]{9.15s}&\makecell[c]{76.25\%}&\makecell[c]{1340.05}&\makecell[c]{6060.06}&\makecell[c]{658.01s}&\makecell[c]{78.79\%}&\makecell[c]{1288.26}&\makecell[c]{5847.32}&\makecell[c]{61.11s}&\makecell[c]{81.72\%}&\makecell[c]{1195.50}&\makecell[c]{5078.05}&\makecell[c]{58.09\%}&\makecell[c]{2151.00}&\makecell[c]{8407.54}\cr\hline
    DS3&15000&\textbf{\makecell[c]{2.27s}}&\textbf{\makecell[c]{84.93\%}}&\textbf{\makecell[c]{1595.60}}&\textbf{\makecell[c]{6918.30}}&\makecell[c]{19.65s}&\makecell[c]{76.85\%}&\makecell[c]{1980.40}&\makecell[c]{8991.78}&\makecell[c]{1556.65s}&\makecell[c]{78.12\%}&\makecell[c]{1949.94}&\makecell[c]{8848.67}&\makecell[c]{95.70s}&\makecell[c]{81.52\%}&\makecell[c]{1792.27}&\makecell[c]{7624.70}&\makecell[c]{58.06\%}&\makecell[c]{3232.07}&\makecell[c]{12625.95}\cr\hline
    DS4&20000&\textbf{\makecell[c]{3.02s}}&\textbf{\makecell[c]{85.27\%}}&\textbf{\makecell[c]{2114.13}}&\textbf{\makecell[c]{9174.47}}&\makecell[c]{35.63s}&\makecell[c]{77.04\%}&\makecell[c]{2618.80}&\makecell[c]{11938.36}&\makecell[c]{3244.99s}&\makecell[c]{78.48\%}&\makecell[c]{2611.09}&\makecell[c]{11767.78}&\makecell[c]{127.41s}&\makecell[c]{82.24\%}&\makecell[c]{2374.07}&\makecell[c]{10092.80}&\makecell[c]{58.31\%}&\makecell[c]{4291.30}&\makecell[c]{16757.45}\cr\hline
    DS5&25000&\textbf{\makecell[c]{3.81s}}&\textbf{\makecell[c]{84.99\%}}&\textbf{\makecell[c]{2661.26}}&\textbf{\makecell[c]{11540.25}}&\makecell[c]{46.48s}&\makecell[c]{77.98\%}&\makecell[c]{3223.50}&\makecell[c]{14669.68}&\makecell[c]{5448.58s}&\makecell[c]{78.27\%}&\makecell[c]{3233.81}&\makecell[c]{14650.77}&\makecell[c]{155.82s}&\makecell[c]{81.70\%}&\makecell[c]{2965.7}&\makecell[c]{12613.45}&\makecell[c]{57.99\%}&\makecell[c]{5367.63}&\makecell[c]{20963.20}\cr\hline
    DS6&30000&\textbf{\makecell[c]{4.65s}}&\textbf{\makecell[c]{85.31\%}}&\textbf{\makecell[c]{3211.40}}&\textbf{\makecell[c]{13926.30}}&\makecell[c]{68.30s}&\makecell[c]{77.22\%}&\makecell[c]{3951.57}&\makecell[c]{17928.70}&\makecell[c]{10932.67s}&\makecell[c]{78.22\%}&\makecell[c]{3930.80}&\makecell[c]{17750.57}&\makecell[c]{194.67s}&\makecell[c]{82.09\%}&\makecell[c]{3568.77}&\makecell[c]{15175.54}&\makecell[c]{58.15\%}&\makecell[c]{6462.23}&\makecell[c]{25242.30}\cr\hline
    DS7&35000&\textbf{\makecell[c]{5.42s}}&\textbf{\makecell[c]{85.11\%}}&\textbf{\makecell[c]{3693.43}}&\textbf{\makecell[c]{16026.43}}&\makecell[c]{79.76s}&\makecell[c]{77.68\%}&\makecell[c]{4585.46}&\makecell[c]{20809.70}&\makecell[c]{19514.45s}&\makecell[c]{78.20\%}&\makecell[c]{4571.50}&\makecell[c]{20676.42}&\makecell[c]{228.03s}&\makecell[c]{82.01\%}&\makecell[c]{4169.03}&\makecell[c]{17730.90}&\makecell[c]{58.18\%}&\makecell[c]{7544.03}&\makecell[c]{29464.20}\cr\hline
    DS8&40000&\textbf{\makecell[c]{6.25s}}&\textbf{\makecell[c]{85.63\%}}&\textbf{\makecell[c]{4281.73}}&\textbf{\makecell[c]{18578.08}}&\makecell[c]{83.55s}&\makecell[c]{78.36\%}&\makecell[c]{5174.06}&\makecell[c]{23597.70}&\makecell[c]{N/A}&\makecell[c]{NA}&\makecell[c]{N/A}&\makecell[c]{N/A}&\makecell[c]{261.61s}&\makecell[c]{82.02\%}&\makecell[c]{4783.83}&\makecell[c]{20346.30}&\makecell[c]{58.14\%}&\makecell[c]{8658.83}&\makecell[c]{33812.69}\cr\hline
    DS9&45000&\textbf{\makecell[c]{6.94s}}&\textbf{\makecell[c]{85.42\%}}&\textbf{\makecell[c]{4759.33}}&\textbf{\makecell[c]{20803.48}}&\makecell[c]{107.52s}&\makecell[c]{77.76\%}&\makecell[c]{5872.37}&\makecell[c]{26718.80}&\makecell[c]{N/A}&\makecell[c]{N/A}&\makecell[c]{N/A}&\makecell[c]{N/A}&\makecell[c]{303.62s}&\makecell[c]{81.92\%}&\makecell[c]{5371.1}&\makecell[c]{22842.35}&\makecell[c]{58.11\%}&\makecell[c]{9720.33}&\makecell[c]{37960.60}\cr\hline
    DS10&50000&\textbf{\makecell[c]{7.66s}}&\textbf{\makecell[c]{85.01\%}}&\textbf{\makecell[c]{5215.50}}&\textbf{\makecell[c]{22620.94}}&\makecell[c]{191.35s}&\makecell[c]{76.04\%}&\makecell[c]{6662.77}&\makecell[c]{30265.15}&\makecell[c]{N/A}&\makecell[c]{N/A}&\makecell[c]{N/A}&\makecell[c]{N/A}&\makecell[c]{332.79s}&\makecell[c]{81.94\%}&\makecell[c]{5931.03}&\makecell[c]{25232.38}&\makecell[c]{58.14\%}&\makecell[c]{10747.77}&\makecell[c]{41975.6}\cr
    \bottomrule
    \end{tabular}
    \end{threeparttable}
    }
\end{table*}

The experiments results, in terms of overall the performance indicators mentioned in subsection V.C, are obtained by all compared algorithms in 30 independent operations and tabulated in Table \ref{t3}. The superior average experimental results are represented by bold. Since OEMACS has reached 19514.45s running time in test data set DS7, which has few practical application value. The experimental effect of OEMACS in test data sets DS8-DS10 are not given in Table \ref{t3}. The running time of FFD is only few micro seconds in all test data set, so it is not listed.

\begin{figure}
\captionsetup{font={footnotesize}}
\centering
\includegraphics[scale=0.32]{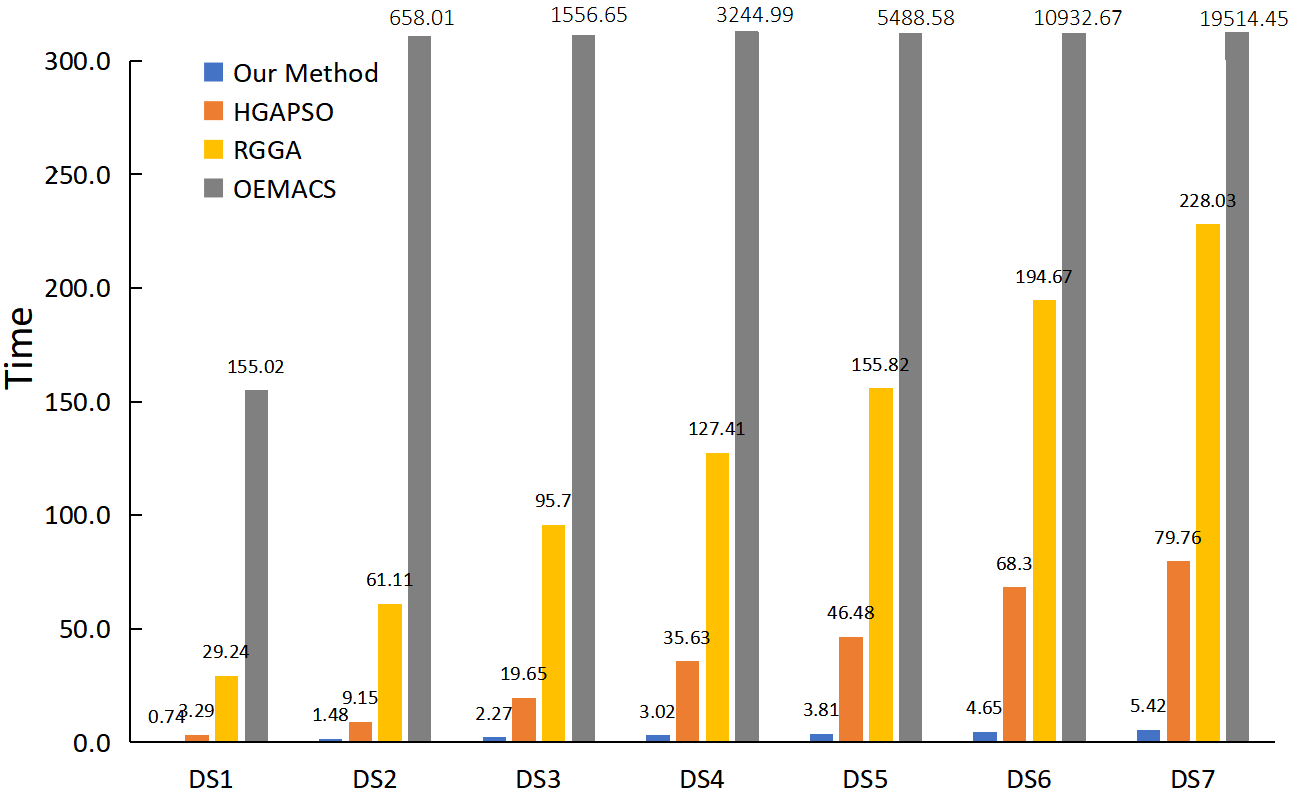}
\captionsetup{justification=centering}
\caption{The running time of all compared algorithm on the DS1-DS7}
\label{f5}
\end{figure}

It can be seen from Table \ref{t3} that on all the test data sets, the proposed method achieves significant superior running time compared to other heuristic algorithms. Explicitly, on the test data set DS7, although the fitness evaluation number of our proposed method is 12250, it completes the task of positioning and optimizing the VMs list used only 5.42s, which proves that the proposed method has favorable real-time capability and expansibility. The reason is that the representation of individual in our method is more effective than that of other heuristic algorithms, which result in dramatically shorten the running time. Compared with the other two modified genetic algorithms HGAPSO and RGGA, they run 79.76s and 228.03s on the test data set DS7, respectively. It must be pointed out that HGAPSO can achieve less running time than RGGA due to the original parameter setting of HGAPSO, in which the number of fitness evaluation is much less than that of RGGA. OEMACS need 19514.45s to complete the assignment process on the DS7, which is the most longest running time in all compared algorithms. Further, it can be seen from tendency Fig. \ref{f5} of the running time increase is that the proposed method adds only about 1.0s to each additional 5000 VMs list, while other algorithms increase the larger steps. Particularly the increase in running time of OEMACS is the most obvious. The experimental result in terms of the running time proves that the proposed method is more suitable for solving the VMP problem in the large-scale data center.

\begin{figure}
\captionsetup{font={footnotesize}}
\centering
\includegraphics[scale=0.35]{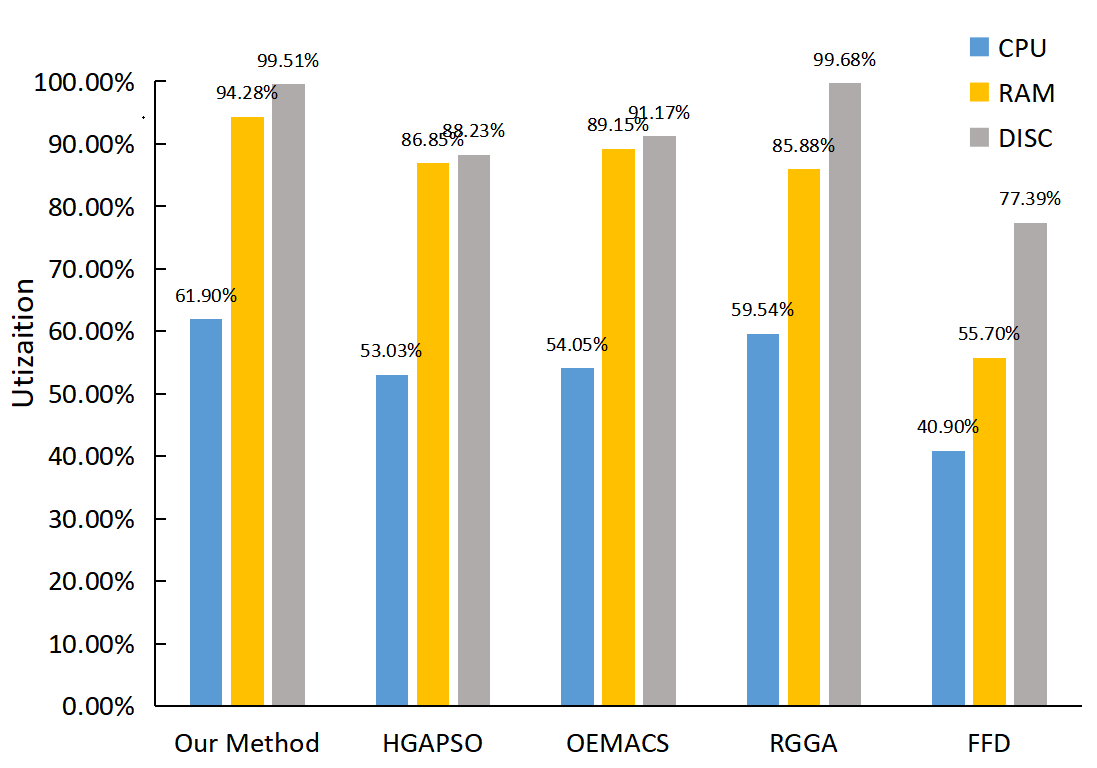}
\captionsetup{justification=centering}
\tiny
\caption{ The average utilization of CPU, RAM and disk of all active servers on the DS7}
\label{f6}
\end{figure}

In terms of the average comprehensive utilization rate of activated PSs cluster, our method has achieved the highest comprehensive utilization rate among all the compared algorithms, which shows that it can effectively maximize the utilization rate of the activated PSs cluster. In a deeper matter, Fig. \ref{f6} shows the utilization rate of CPU, RAM, and disk on the DS7 of all compared algorithms, respectively. As shown in Fig. \ref{f6}, compared to RGGA, the proposed method obtains the superior utilization rate on CPU and RAM, and comparable utilization rate on disk, which is considered as bottleneck resource in this experiment. And the proposed method shows higher utilization rate on CPU, RAM and disk in the comparison of HGAPSO, OEMACS and FFD. It shows that our method could effectively respond to LVMP problems of which has the bottleneck resource characteristics.

In addition, the average activated number of PSs cluster by the proposed method is less than other compared heuristic algorithm on all test data sets. For example, the average activated number of PSs in the proposed method is 3693.43 on the DS7, which HGAPSO, OEMACS and RGGA are 4585.46, 4571.50 and 4169.03 respectively. The results proves that our method can effectively reduce the number of activated PSs, accordingly achieve the purpose of reducing energy consumption.

Finally, compared with other algorithms, it can be seen from Table \ref{t3} that the proposed method obtains lower deployment cost than other compared algorithm. This is due to the fact that each VMP optimization task is modeled as an optimization goal with reduced deployment costs in our proposed method. The advantage of cost based model is its ability to lower the deployment cost of data center and deal with VMP problems in heterogeneous environments. Moreover, since each VMP optimization task has the same optimization goals, which means that there are some higher similarity between them, it can lead to effective communication and obtain better optimization results.

\subsection{Parametric Analysis}

\begin{figure}
\captionsetup{font={footnotesize}}
\centering
\includegraphics[scale=0.5]{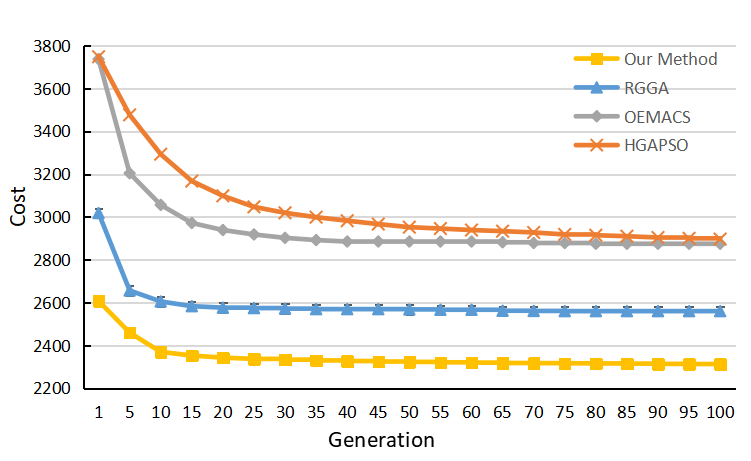}
\captionsetup{justification=centering}
\caption{ Convergence test of our proposed method on the DS1}
\label{f9}
\end{figure}

In this experiment, the susceptibility of two parameters in the proposed method is tested on the DS1 independently, e.g. random mating probability \emph{rmp} and the number of VMs assigned to each task $N_i$. The situations on the other test data sets are similar.

\begin{figure}
\centering
\includegraphics[scale=0.48]{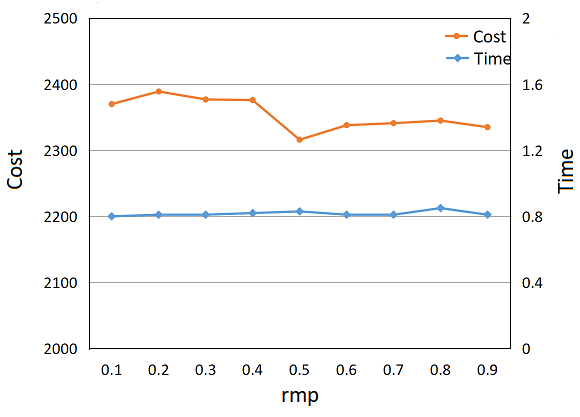}
\captionsetup{font={footnotesize}}
\caption{The effect of different value of \emph{rmp} on the DS1}
\label{f7}
\end{figure}

To analysis the susceptibility of \emph{rmp}, we set its value from 0.1 to 0.9 with a step length of 0.1, and the experiment results are displayed in Fig. \ref{f7}. It could be seen from Fig. \ref{f7} that as the value of \emph{rmp} increases, the running times of our proposed method are almost the same, which proves that the change in \emph{rmp} value has little effect on the running times. In addition, the proposed method obtains the superior experiment result in terms of deployment cost when \emph{rmp} = 0.5. The reason can be summarized as follow. The same deployment cost based objective optimization function is used for all VMP optimization tasks. It indicates that the knowledge exchange between tasks is positive at a high probability during the process of evolution. And the larger value of \emph{rmp} increases the probability of knowledge exchange between tasks, then obtains a better result. However, as the value of \emph{rmp} continues to grow, the deployment cost may be gradually increased. The reason is that the larger value of \emph{rmp} may cause the sub-population specify to the tasks to fall into local optimum easily due to the loss of population diversity. In summary, the \emph{rmp} is set to \emph{rmp} = 0.5.

\begin{figure}
\captionsetup{font={footnotesize}}
\centering
\includegraphics[scale=0.45]{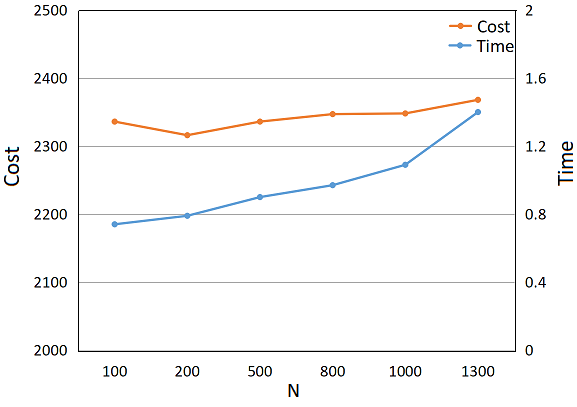}
\captionsetup{justification=centering}
\caption{ The effect of the number of VMs assigned to each task on the DS1}
\label{f8}
\end{figure}

To test the effect of the number of $N_i$, making $N_i$ = 100, 200, 500, 800, 1000, 1300, 1500, respectively, and the results are shown in Fig. \ref{f8}. It can be seen from Fig \ref{f8} that, as the value of $N_i$ increases, the running time is increased and the deployment cost has also risen. This is due to the fact that the increase of $N_i$ expand the search space of the sub-population from different tasks, resulting in the quality of the solution decreased. It is worth noting that the $N_i$ is not the smaller, the better. Although the smaller $N_i$ can bring advantages in the representation of individuals, the search space of the sub-population specify to the tasks becomes narrow, and the sub-population is easy to fall into local optimum. Summarily the $N_i$ was set to $N_i$ = 200.

\subsection{Convergence Analysis}

Take DS1 as an example, the convergence analysis of our proposed method is presented as follows. The situations on the other test data sets are similar. The maximum number of iterations is set as 100 in this experiment. Fig. \ref{f9} depicts the convergence curves of the proposed method and other compared algorithms. It can be seen from Fig. \ref{f9} that our method obtains better results than other compared algorithms in the first generation, which is crucial to guide the search process. And the convergence of our method is slowing down at the tenth iteration and approaching the local optima with deployment cost 2388.86. Actually, at the early iteration, the proposed method has converged to a superior local optima solution than other compared algorithms. The reason can be summarized as follows. On the one hand, the greedy-based allocation operator is competent to finish the placement of VMs onto PSs, which is able to find superior local optimal solutions of the VMP optimization task at the early iteration. On the other hand, owing to the more effective representation block of the individual in the proposed method, the search space of each VMP optimization task is narrower and it is conducive to finding a local optima solution. The above results show that the proposed method can converge to a better local optima solution more fastly.

\subsection{Single-Factorial Solver vs Multi-Factorial Solver}
In order to explore the effectiveness of multi-factorial optimization technology, the multi-factorial method proposed in this paper is modified to the single-factorial version (SFEA) to compare. In this experiment, SFEA adopts the same initialization operator, crossover and mutation operator, greedy-based allocation operator and environment selection operator as the proposed method. The difference between itself and the proposed multi-factorial method is that SFEA lacks the process of establishing a unified express space. Meanwhile, SFEA is a single-factorial solver and it has no inter-task communication. For fair comparison, the same parameters as the proposed multi-factorial method are used in SFEA. The test data sets DS1 to DS10 are conducted to assess the performance of SFEA, which the experiment result are summarized in Table \ref{t4}.

As seen in Table \ref{t4}, the running times of SFEA on all test instances are longer than its multi-factorial version. This is attributed to the fact that the representation block of individuals in the multi-factorial environment can not only decode into its related task, but also retain the genetic information of other tasks, which resulting in its effective. In the other three performance indicator (comprehensive utilization rate, the number of activated PSs and deployment cost), the proposed multi-factorial method obtains superior results than SFEA on all the test data sets. The reason can be summarized as follows. The search space of population in SFEA is large and complex, which causes slower convergence speed of SFEA. Reversely, the search space of population in the multi-factorial version is relatively narrow, so it can converge quickly. Besides these, the population of the multi-factorial version is not easy to fall into local optimum due to the knowledge transfer appeared among tasks, which is also the key to obtain superior solutions than SFEA.

\renewcommand{\arraystretch}{0.7} 
\linespread{1.2}
\begin{table}[tp]
  \setlength{\tabcolsep}{1mm}{
  \fontsize{5.7}{9}\selectfont
  \begin{threeparttable}
  \captionsetup{font={small}}
  \caption{EXPERIMENTAL RESULT COMPARISONS WITH DIFFERENT SIZE OF VMs ON THE PROPOSED METHOD AND SFEA}
  \label{t4}
    \begin{tabular}{cccccccccc}
    \toprule
    \multirow{1}{*}{Data Set NO.}&\multirow{1}{*}{Size}&
    \multicolumn{4}{c}{Our Method}&\multicolumn{4}{c}{SFEA}\cr
    \cmidrule(lr){3-6}\cmidrule(lr){7-10}
    &&Time&Util&Num&Cost&Time&Util&Num&Cost\cr
    \midrule
    \renewcommand{\arraystretch}{2}
    DS1&5000&\textbf{\makecell[c]{0.74s}}&\textbf{\makecell[c]{85.47\%}}&\textbf{\makecell[c]{534.26}}&\textbf{\makecell[c]{2316.15}}&\makecell[c]{2.94s}&\makecell[c]{83.60\%}&\makecell[c]{565.13}&\makecell[c]{2430.31}\cr\hline
    DS2&10000&\textbf{\makecell[c]{1.48s}}&\textbf{\makecell[c]{85.19\%}}&\textbf{\makecell[c]{1056.01}}&\textbf{\makecell[c]{4578.87}}&\makecell[c]{10.62s}&\makecell[c]{82.93\%}&\makecell[c]{1130.80}&\makecell[c]{4862.40}\cr\hline
    DS3&15000&\textbf{\makecell[c]{2.27s}}&\textbf{\makecell[c]{84.93\%}}&\textbf{\makecell[c]{1595.60}}&\textbf{\makecell[c]{6918.30}}&\makecell[c]{27.21s}&\makecell[c]{82.71\%}&\makecell[c]{1697.20}&\makecell[c]{7311.37}\cr\hline
    DS4&20000&\textbf{\makecell[c]{3.02s}}&\textbf{\makecell[c]{85.27\%}}&\textbf{\makecell[c]{2114.13}}&\textbf{\makecell[c]{9174.47}}&\makecell[c]{52.71s}&\makecell[c]{83.24\%}&\makecell[c]{2259.13}&\makecell[c]{9722.77}\cr\hline
    DS5&25000&\textbf{\makecell[c]{3.81s}}&\textbf{\makecell[c]{84.99\%}}&\textbf{\makecell[c]{2661.26}}&\textbf{\makecell[c]{11540.25}}&\makecell[c]{85.01s}&\makecell[c]{82.76\%}&\makecell[c]{2812.80}&\makecell[c]{12115.50}\cr\hline
    DS6&30000&\textbf{\makecell[c]{4.65s}}&\textbf{\makecell[c]{85.31\%}}&\textbf{\makecell[c]{3211.40}}&\textbf{\makecell[c]{13926.30}}&\makecell[c]{117.71s}&\makecell[c]{82.84\%}&\makecell[c]{3407.10}&\makecell[c]{14652.40}\cr\hline
    DS7&35000&\textbf{\makecell[c]{5.42s}}&\textbf{\makecell[c]{85.11\%}}&\textbf{\makecell[c]{3693.43}}&\textbf{\makecell[c]{16026.43}}&\makecell[c]{154.72s}&\makecell[c]{82.72\%}&\makecell[c]{3982.33}&\makecell[c]{17132.02}\cr\hline
    DS8&40000&\textbf{\makecell[c]{6.25s}}&\textbf{\makecell[c]{85.63\%}}&\textbf{\makecell[c]{4281.73}}&\textbf{\makecell[c]{18578.08}}&\makecell[c]{201.69s}&\makecell[c]{82.90\%}&\makecell[c]{4564.73}&\makecell[c]{19636.10}\cr\hline
    DS
    9&45000&\textbf{\makecell[c]{6.94s}}&\textbf{\makecell[c]{85.42\%}}&\textbf{\makecell[c]{4759.33}}&\textbf{\makecell[c]{20803.48}}&\makecell[c]{255.74s}&\makecell[c]{82.71\%}&\makecell[c]{5129.00}&\makecell[c]{22057.97}\cr\hline
    DS10&50000&\textbf{\makecell[c]{7.66s}}&\textbf{\makecell[c]{85.01\%}}&\textbf{\makecell[c]{5215.50}}&\textbf{\makecell[c]{22620.94}}&\makecell[c]{318.21s}&\makecell[c]{82.60\%}&\makecell[c]{5667.50}&\makecell[c]{24376.51}\cr
    \bottomrule
    \end{tabular}
    \end{threeparttable}
    }
\end{table}

\subsection{Extra Large-scale Test Data Sets}
In this experiment, to further assess the performance of our proposed method on extra large-scale VMP problems, the extra test data sets with a maximum size of 250,000 VMs is created as the same approach of the subsection V.A. Since OEMACS and FFD have been proved to be unsuitable for solving LVMP problems in the above experiments. In this experiment, the performance of our method is assessed in comparison with HGAPSO and RGGA. The experiment results are summarized in Table \ref{t5}. As can be seen from Table \ref{t5}, on the EDS6, the proposed method completes the assignment process for only 37.61s. The experiment results, in terms of the average comprehensive utilization rate, deployment cost and the average number of activated PSs, the proposed method also shows superior performance than HGAPSO and RGGA on all the extra test data sets.

\begin{figure}
\centering
\includegraphics[scale=0.3]{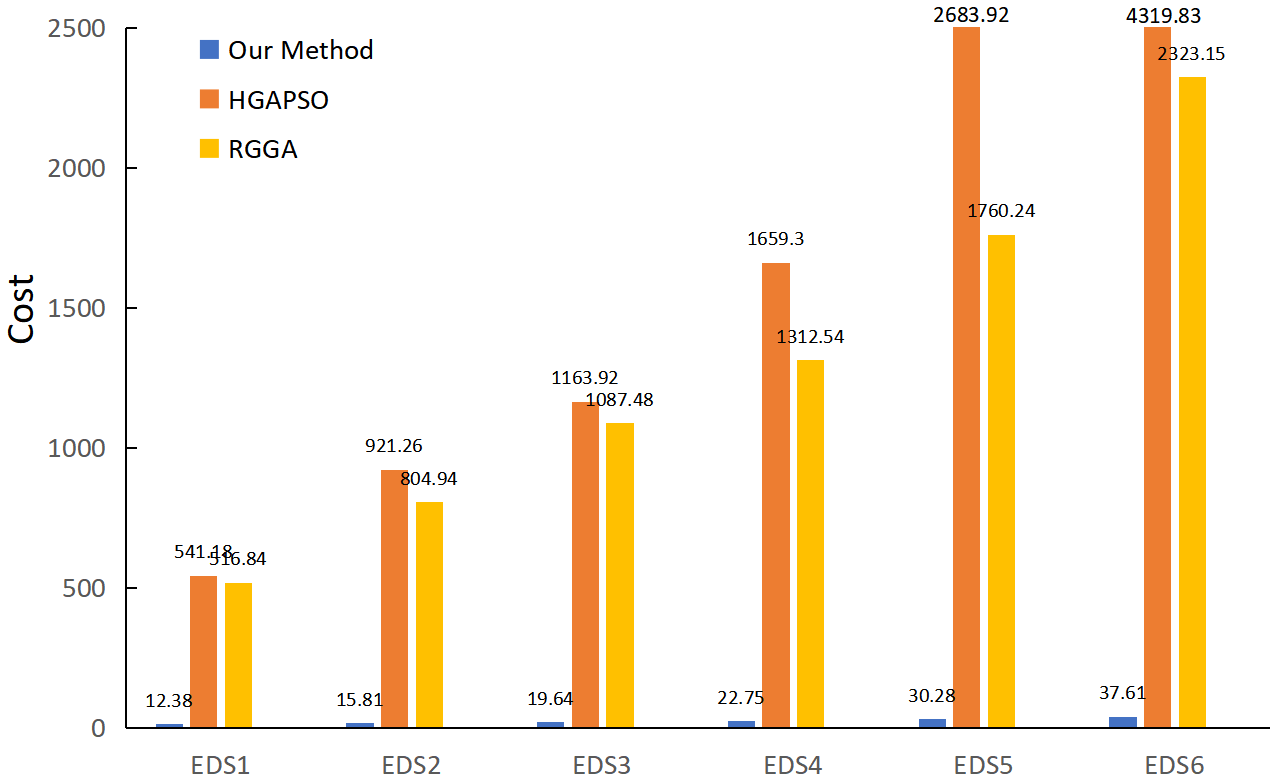}
\captionsetup{justification=centering}
\captionsetup{font={footnotesize}}
\caption{The average running time of the proposed method, HGAPSO and RGGA on the EDS1-EDS6}
\label{f10}
\end{figure}

Fig. \ref{f10} shows the average running time of the proposed method and compared algorithm on the extra large-scale test data sets. It can be seen from Fig. \ref{f10}, as the test data sets increases at a large step, the proposed method has the minimum and stability step of increased time, with the running time of only about 7s per 50,000 VMs added. The steps of HGAPSO and RGGA about increased time are relatively large, and their increased time step has a gradually increased tendency. The tuning time of HGAPSO is even longer on all the extra test data sets than RGGA. The experimental results proves that the proposed method can be really applied to deal with the problem of extra large-scale VMP problems, and has the good expansibility and real-time performance.

\renewcommand{\arraystretch}{0.7} 
\linespread{1.5}
\begin{table*}[tp]
  \centering
  \setlength{\tabcolsep}{1mm}{
  \fontsize{6}{9}\selectfont
  \begin{threeparttable}
  \captionsetup{font={footnotesize}}
  \caption{EXPERIMENTAL RESULT OF EXTRA LARGE-SCALE VMs Test Data Sets}
  \label{t5}
    \begin{tabular}{cccccccccccccc}
    \toprule
    \multirow{1}{*}{Extra Data Set NO.}&\multirow{1}{*}{Size}&
    \multicolumn{4}{c}{Our Method}&\multicolumn{4}{c}{HGAPSO}&\multicolumn{4}{c}{RGGA}\cr
    \cmidrule(lr){3-6}\cmidrule(lr){7-10}\cmidrule(lr){11-14}
    &&Time&Util&Num&Cost&Time&Util&Num&Cost&Time&Util&Num&Cost\cr
    \midrule
    \renewcommand{\arraystretch}{2}
    EDS1&80000&\textbf{\makecell[c]{12.38s}}&\textbf{\makecell[c]{84.92\%}}&\textbf{\makecell[c]{8516.60}}&\textbf{\makecell[c]{36915.98}}&\makecell[c]{541.18s}&\makecell[c]{75.82\%}&\makecell[c]{10786.20}&\makecell[c]{4875.21}&\makecell[c]{516.84s}&\makecell[c]{81.98\%}&\makecell[c]{9475.32}&\makecell[c]{40367.05}\cr\hline
    EDS2&100000&\textbf{\makecell[c]{15.81s}}&\textbf{\makecell[c]{85.06\%}}&\textbf{\makecell[c]{11615.23}}&\textbf{\makecell[c]{46057.90}}&\makecell[c]{921.26s}&\makecell[c]{77.62\%}&\makecell[c]{13057.87}&\makecell[c]{59372.83}&\makecell[c]{804.94s}&\makecell[c]{82.01\%}&\makecell[c]{11885.54}&\makecell[c]{50558.84}\cr\hline
    EDS3&130000&\textbf{\makecell[c]{19.64s}}&\textbf{\makecell[c]{84.42\%}}&\textbf{\makecell[c]{13780.86}}&\textbf{\makecell[c]{59762.06}}&\makecell[c]{1163.92s}&\makecell[c]{78.99\%}&\makecell[c]{16680.00}&\makecell[c]{75950.89}&\makecell[c]{1087.48}&\makecell[c]{82.10\%}&\makecell[c]{15498.02}&\makecell[c]{65898.27}\cr\hline
    EDS4&150000&\textbf{\makecell[c]{22.75s}}&\textbf{\makecell[c]{85.76\%}}&\textbf{\makecell[c]{16054.73}}&\textbf{\makecell[c]{69619.23}}&\makecell[c]{1659.30s}&\makecell[c]{76.78\%}&\makecell[c]{19801.67}&\makecell[c]{89993.44}&\makecell[c]{1312.54s}&\makecell[c]{82.15\%}&\makecell[c]{17741.20}&\makecell[c]{75623.81}\cr\hline
    EDS5&200000&\textbf{\makecell[c]{30.28s}}&\textbf{\makecell[c]{84.88\%}}&\textbf{\makecell[c]{21062.23}}&\textbf{\makecell[c]{91355.30}}&\makecell[c]{2683.92s}&\makecell[c]{75.80\%}&\makecell[c]{27005.57}&\makecell[c]{122151.32}&\makecell[c]{1760.24s}&\makecell[c]{82.22\%}&\makecell[c]{23521.56}&\makecell[c]{100441.81}\cr\hline
    EDS6&250000&\textbf{\makecell[c]{37.61s}}&\textbf{\makecell[c]{85.65\%}}&\textbf{\makecell[c]{26453.77}}&\textbf{\makecell[c]{114808.82}}&\makecell[c]{4319.83s}&\makecell[c]{73.72\%}&\makecell[c]{34277.91}&\makecell[c]{154519.62}&\makecell[c]{2323.15s}&\makecell[c]{82.20\%}&\makecell[c]{29561.23}&\makecell[c]{125991.47}\cr
    \bottomrule
    \end{tabular}
    \end{threeparttable}
    }
\end{table*}

\section{Conclusion and Future Work}

This paper provides a MFO method to complete the placement of VMs onto PSs in heterogeneous environments of the large-scale data center. We firstly reformulate a deployment cost based VMP problem in the form of the MFO problem. Multiple VMP optimization tasks based on deployment cost are modeled to achieve an organic composition of different configuration of PSs in heterogeneous environments, which lead to reduces the deployment costs directly for cloud providers. Furthermore, a multi-factorial evolutionary algorithm coupled with the greedy-based allocation operator is proposed to address the established MFO problem. After that, a re-migration and merge operator is designed to obtain the integrated solution of the LVMP problem. The comprehensive experiments, including the large-scale and extra large-scale test data sets, are conducted to assess the effectiveness of our proposed method. The experimental result describes that the proposed method can significantly lower the optimization time and provide an effective placement solution for VMP problem in the large-scale data center.

Although our proposed method shows promising problem-solving for the LVMP problems in heterogeneous environments, it only considered the optimization target of resource allocation. The actual data center is faced not only with the challenge in resource scheduling optimization but also included other complex optimization objectives, such as energy consumption optimization, network traffic optimization and so on. For future work, how to effectively integrate many different optimization objectives will continue to be explored based on MFO technology.

\begin{spacing}{1}
    \small
    \bibliographystyle{IEEEtran}
    \bibliography{MFO}
\end{spacing}
\end{document}